\documentclass{article} 
\usepackage{iclr2021_conference,times}

\usepackage{amsmath,amsfonts,bm}

\def\eqref#1{equation~\ref{#1}}

\def\1{\bm{1}}

\DeclareMathAlphabet{\mathsfit}{\encodingdefault}{\sfdefault}{m}{sl}
\SetMathAlphabet{\mathsfit}{bold}{\encodingdefault}{\sfdefault}{bx}{n}

\newcommand{\E}{\mathbb{E}}
\newcommand{\Ls}{\mathcal{L}}

\usepackage{hyperref}
\usepackage{url}

\usepackage{amssymb}
\usepackage{mathtools}
\usepackage{subcaption}
\usepackage{multirow}
\usepackage{adjustbox}
\usepackage{xcolor}
\usepackage{enumitem}
\usepackage{wrapfig}

\newcommand{\Arch}{\mathcal{A}}
\newcommand{\Net}{\mathcal{N}}

\title{How does Weight Sharing Help in \\ Neural Architecture Search?}

\author{Yuge Zhang, Quanlu Zhang \& Yaming Yang \\
Microsoft Research \\
\texttt{\{Yuge.Zhang,Quanlu.Zhang,Yang.Yaming\}@microsoft.com}
}

\iclrfinalcopy 
\begin{document}

\maketitle

\begin{abstract}
    Weight sharing, as an approach to speed up architecture performance estimation has received wide attention. Instead of training each architecture separately, weight sharing builds a supernet that assembles all the architectures as its submodels. However, there has been debate over whether the NAS process actually benefits from weight sharing, due to the gap between supernet optimization and the objective of NAS. To further understand the effect of weight sharing on NAS, we conduct a comprehensive analysis on five search spaces, including NAS-Bench-101, NAS-Bench-201, DARTS-CIFAR10, DARTS-PTB, and ProxylessNAS. We find that weight sharing works well on some search spaces but fails on others. Taking a step forward, we further identified biases accounting for such phenomenon and the capacity of weight sharing. Our work is expected to inspire future NAS researchers to better leverage the power of weight sharing.
\end{abstract}

\section{Introduction}


Over recent years, weight-sharing NAS has emerged as a way to search for architectures under very limited resources, and has been gaining popularity~\citep{pham2018efficient,liu2018darts,cai2018proxylessnas,bender18understanding}. It builds all architectures into a supernet where each architecture is a submodel. Once the supernet has been trained, architectures can be directly evaluated by inheriting weights from the supernet, thus saving the huge cost to train each architecture from scratch.

Weight sharing is built on the hypothesis that an architecture's performance estimated within a supernet indicates its real performance. However, such hypothesis has been long questioned. Many analytical papers have conducted empirical analysis on whether and how weight sharing helps in finding the best architecture~\citep{bender18understanding,luo2019understanding,yu2019evaluating,yang2019nas,zhang2020deeper,pourchot2020share,yu2020train,bender2020can}. Though valuable insights have been delivered, the understanding of weight sharing still only touches a tip of the iceberg. Futhermore, due to the differences in search spaces and experiment setups, sometimes different works even draw contradictory conclusions~\citep{bender18understanding,li2019random}.

For a deeper understanding, we conduct a more comprehensive study spanning five search spaces: NAS-Bench-101~\citep{ying2019nasbench101}, NAS-Bench-201~\citep{dong2020nasbench201}, DARTS-CIFAR10, DARTS-PTB~\citep{liu2018darts}, and ProxylessNAS~\citep{cai2018proxylessnas}. The study reveals that, for some spaces, NAS can benefit from weight sharing to find a good-performing architecture, while for others, it hardly works. We then analyzed the results to dig out the reason for such phenomenon. Our experiments exclude the factor of sharing density of weights and training instability, and point the cause directly to the bias towards certain operators, connections and model sizes. We noticed that weight sharing performs better when the bias coincides with the actual rank of architectures.

Since weight sharing struggles to identify the best architectures, some works resort to pruning search space in order to increase the probability of finding the best, e.g., RL-based methods~\citep{pham2018efficient,bender2020can} and heuristic methods~\citep{you2020greedynas,hu2020anglebased,chen2019pdarts,li2019improving}. To identify the upper limit to which weight sharing can correctly prune the search space, we compared weight sharing on pruned NAS-Bench series. We came to a conclusion that weight sharing is able to find top-10\% architectures, but struggles to find the best.

To conclude, our contributions can be summarized as follows:

\begin{itemize}[leftmargin=0.3in]
    \item We conduct a comprehensive study on five different search spaces, spanning from CNN to RNN. The pairwise comparison between different spaces reveals more convincing insights, and also explains some contradictory conclusions drawn from previous papers.
    \item To the best of our knowledge, we are the first to systematically analyze the performance of weight sharing on pruned search space.
    \item We released a NAS benchmark containing more than 500 architectures where the ground truth\footnote{Ground truth of a model refers to the performance when it is trained independently and sufficiently.} of each architecture is also reported, which we expect will facilitate NAS research.\footnote{Code and benchmarks are publicly available at \url{https://anonymous.4open.science/r/d6dba7e2-2c9b-48da-9592-2d0bbe41ae7e/}.}
\end{itemize}

\section{Weight Sharing Benchmark}

In this section, we conduct a comprehensive study of weight sharing on five search spaces. For all the experiments, we adopt uniform sampling to train a supernet, i.e., single-path one shot (SPOS)~\citep{guo2019single}, as it is a generalized approach and it can also be seen as a basic component of other advanced sampling methods that exclude other training tricks. Details about experiment setups including search spaces and evaluation methodology can be found in \autoref{sec:experiment-setup}.

\begin{figure}[htbp]
    \centering
    \begin{subfigure}{0.37\textwidth}
        \centering
        \includegraphics[width=\textwidth]{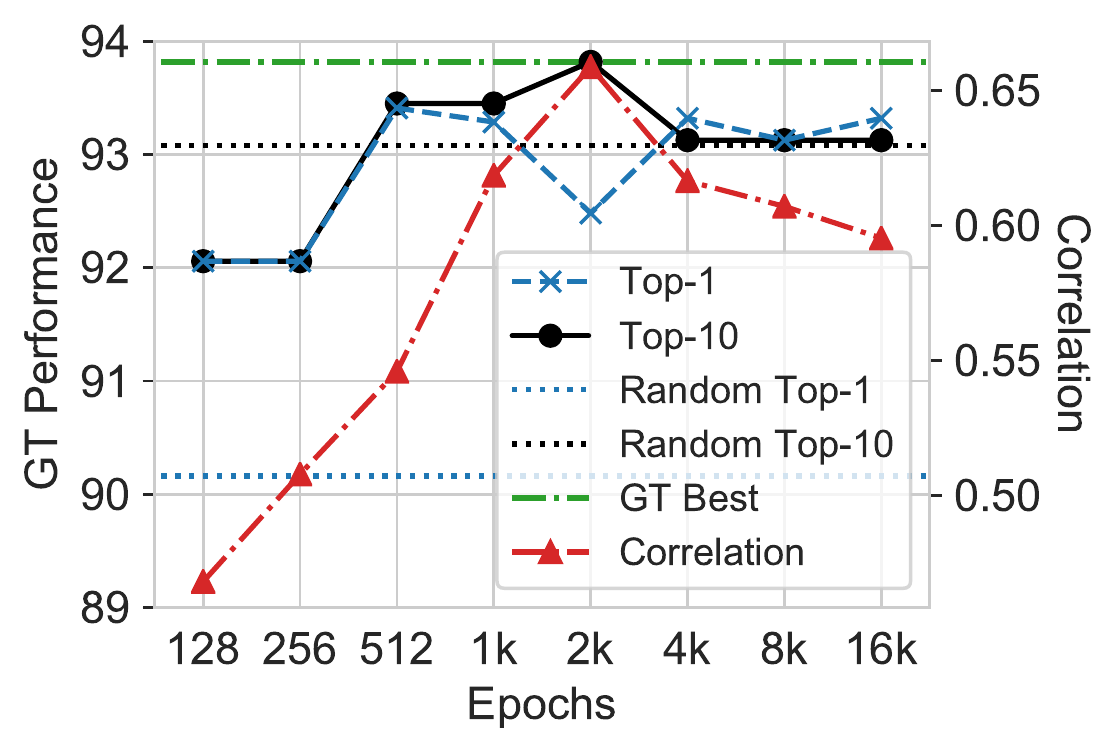}
        \caption{NAS-Bench-101}
    \end{subfigure}
    \begin{subfigure}{0.37\textwidth}
        \centering
        \includegraphics[width=\textwidth]{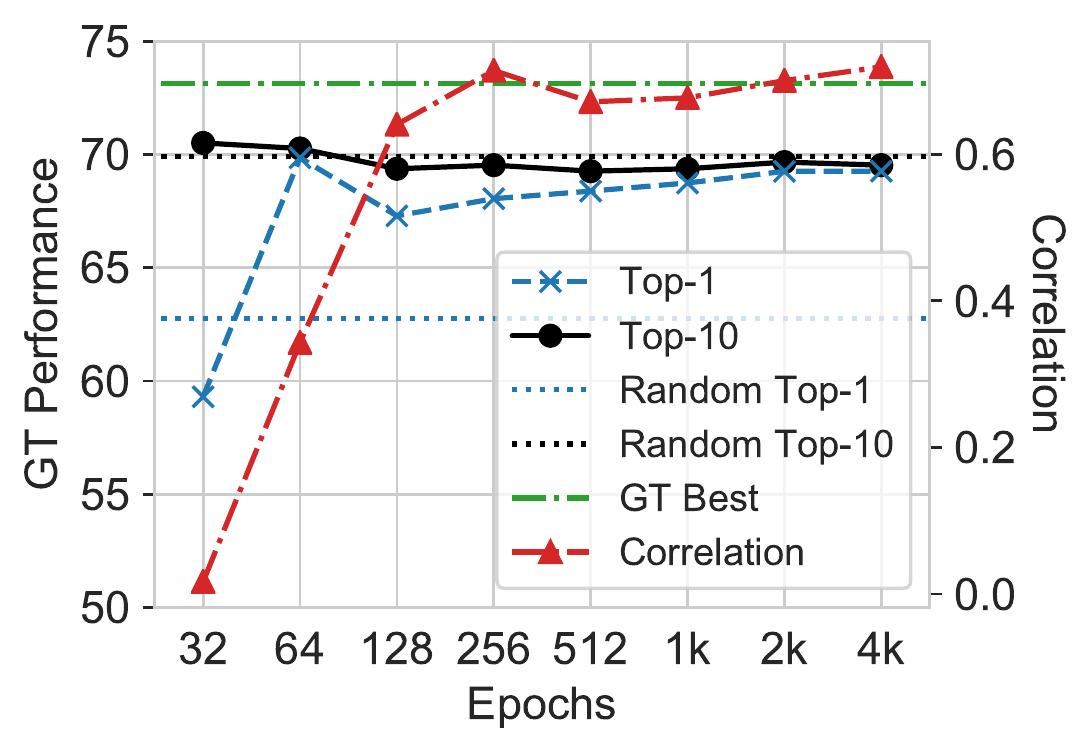}
        \caption{NAS-Bench-201}
    \end{subfigure}
    \begin{subfigure}{0.38\textwidth}
        \centering
        \includegraphics[width=\textwidth]{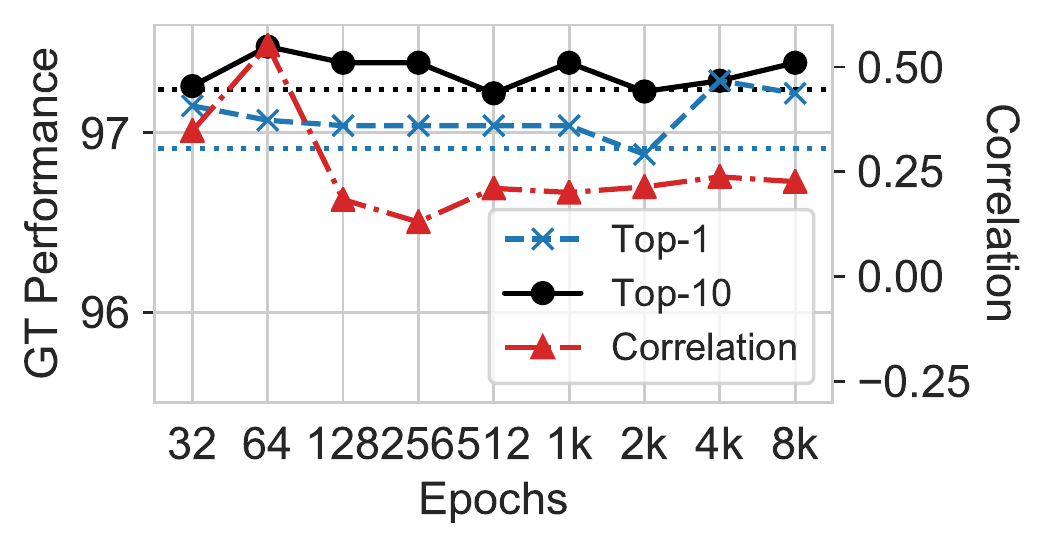}
        \caption{DARTS-CIFAR10}
    \end{subfigure}
    \begin{subfigure}{0.3\textwidth}
        \centering
        \includegraphics[width=\textwidth]{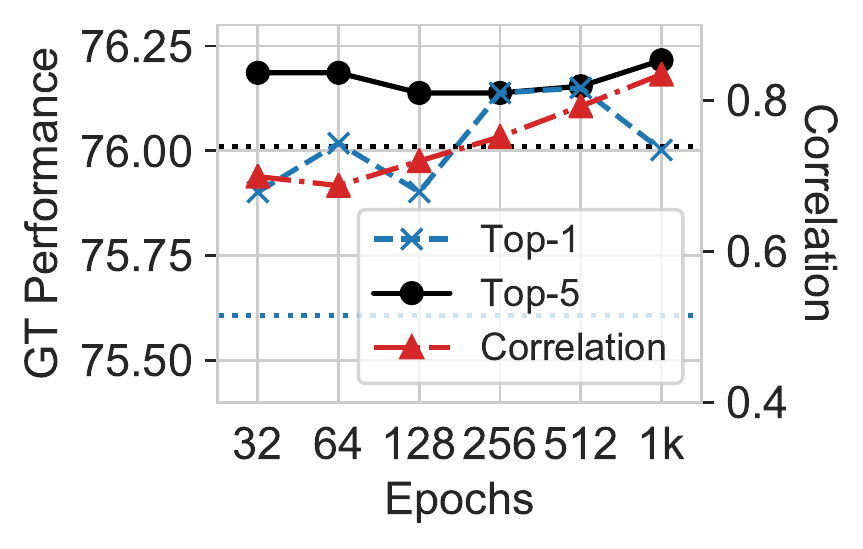}
        \caption{ProxylessNAS}
    \end{subfigure}
    \begin{subfigure}{0.3\textwidth}
        \centering
        \includegraphics[width=\textwidth]{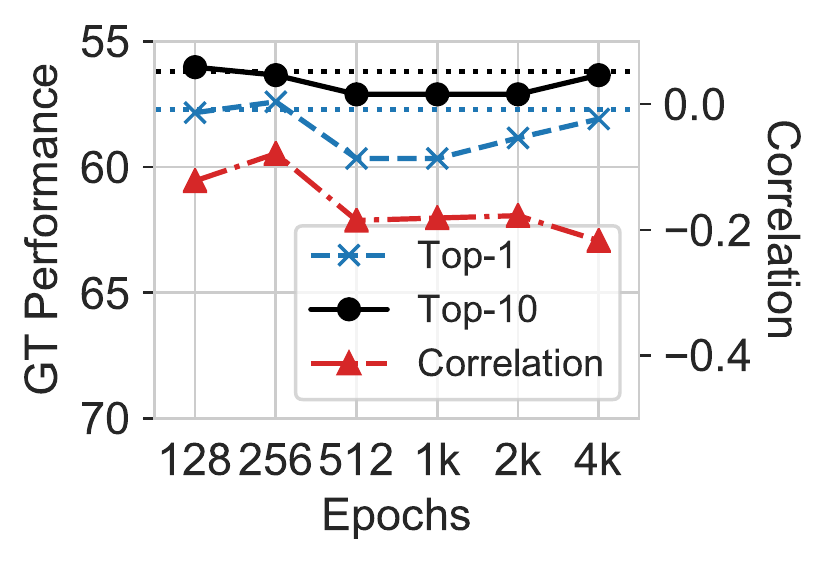}
        \caption{DARTS-PTB}
    \end{subfigure}
    \vspace{-0.5em}
    \caption{We show correlation, top-$k$ (i.e., best ground truth of $k$ architectures selected with the help of weight sharing) and random top-$k$ (i.e., best ground truth of randomly-selected $k$ architectures). We also show the best ground truth (in green dash-dot line) if the metric is available. Details explanation on metrics are in \autoref{sec:experiment-setup}.}
    \label{fig:top10-perf}
    \vspace{-0.5em}
\end{figure}

\noindent \textbf{Longer training does not guarantee better results.} As previous works~\citep{chu2019fairnas,chu2019scarletnas,guo2019single,luo2019understanding} pointed out that supernet needs to be trained sufficiently to benefit NAS, we first vary the training epochs exponentially instead of using a fixed training schedule. From the perspective of average validation performance, supernet converges well and there is no sign of over-fitting (see \autoref{sec:convergence}). However, as shown in \autoref{fig:top10-perf}, a longer training does not guarantee a better ranking model. For example, on DARTS-CIFAR10, correlation achieves 0.54 when trained with 64 epochs, but later decreases to around 0.2 with longer training. On PTB, the correlation can be even worse than 0. There is a similar phenomenon on top-$k$. Although top-$k$ is not very sensitive to the changes in correlation, performance of selected architecture could still drop when supernet is trained longer.

\noindent \textbf{Effectiveness of weight sharing relies on search space.} We draw the conclusion by comparing the results across different search spaces. For example, on NAS-Bench-201, the correlation can be as high as 0.7, whereas using weight-sharing to guide NAS on DARTS-PTB can be even worse than random search. On search spaces where correlation is greater than 0.4, weight-sharing NAS still  outperforms random search by a clear margin on top-$k$ performance.

\begin{wrapfigure}{r}{0.62\textwidth}
    \vspace{-1em}
    \centering
    \begin{subfigure}{0.3\textwidth}
        \centering
        \includegraphics[width=\textwidth]{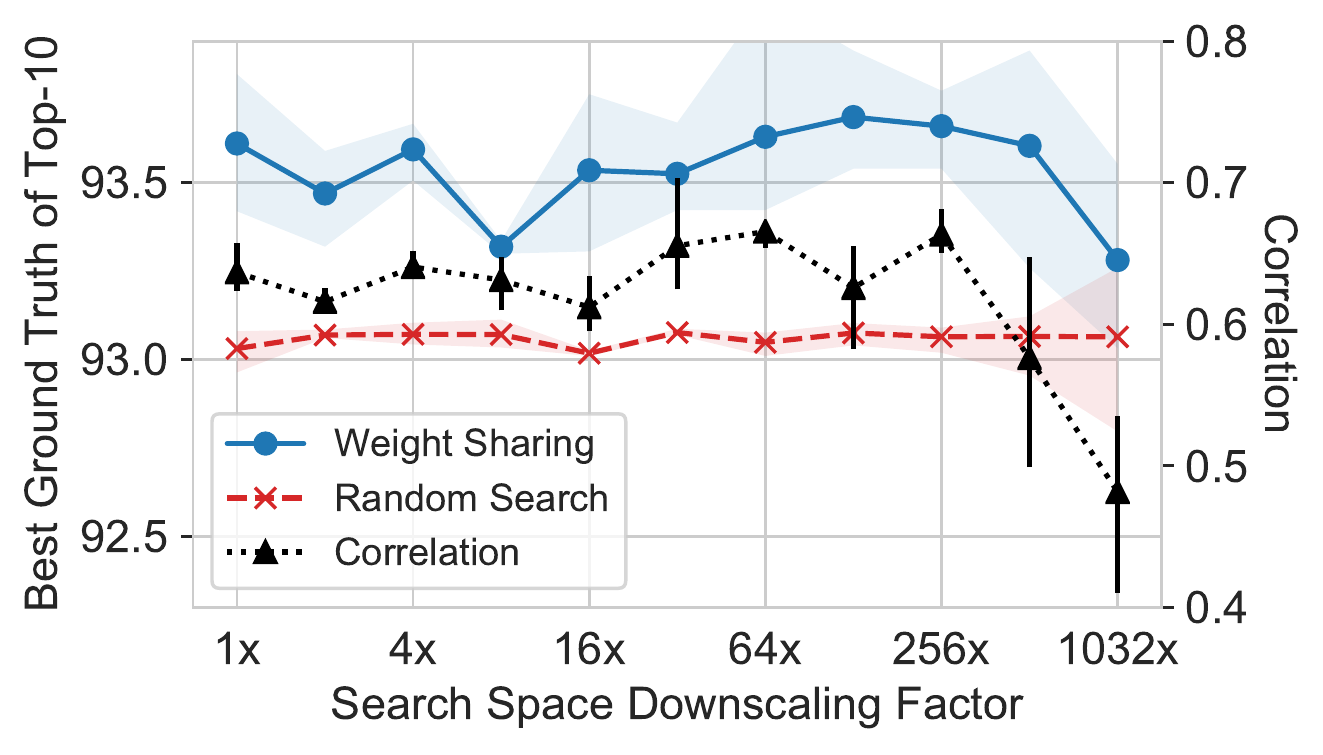}
        \caption{NAS-Bench-101}
    \end{subfigure}
    \begin{subfigure}{0.3\textwidth}
        \centering
        \includegraphics[width=\textwidth]{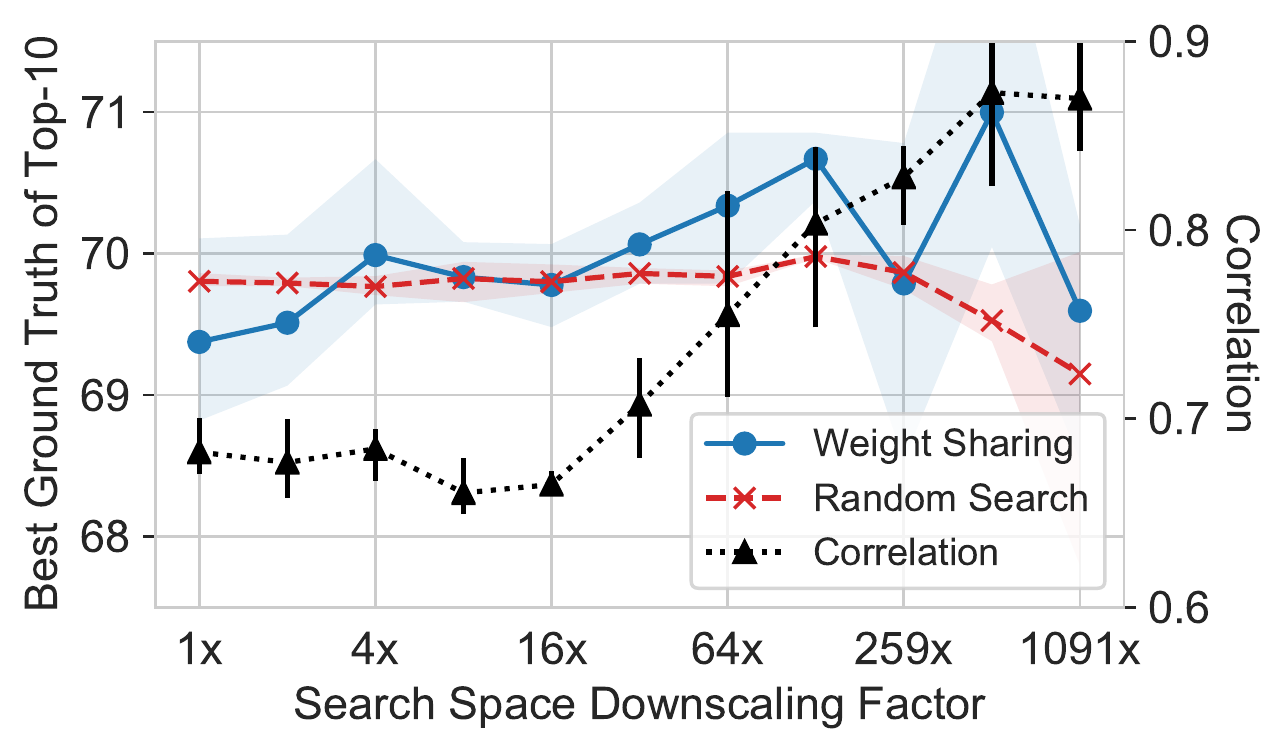}
        \caption{NAS-Bench-201}
    \end{subfigure}
    \caption{Comparison between random search and weight-sharing NAS as we decrease the sharing density by conducting NAS on a randomly-downscaled search space. We show top-10 accuracy (blue and red lines) for weight sharing and random search, respectively, and correlation in black lines, when search space is $k$ times smaller than the original one. The error-bars and regions are min/max across 3 different runs.}
    \label{fig:prune-random}
    \vspace{-1em}
\end{wrapfigure}

\noindent \textbf{Reducing the sharing density hardly helps.}  \cite{zhang2020deeper} pointed out the intuition that submodels have interference with each other, and thus proposed to reduce the density of weight sharing (i.e., how many different architectures are sharing the weight of one operator). However, our experiments show that this is only true when the sharing density is substantially low. For example, in \autoref{fig:prune-random}, the improvement is only noticeable when the downscaling factor is up to 64 on NAS-Bench-201, at which time the search space only contains 244 different architectures. On NAS-Bench-101, no increasing trend is observed at all.

\begin{figure}[htbp]
    \centering
    \begin{subfigure}{0.245\textwidth}
        \centering
        \includegraphics[width=\textwidth]{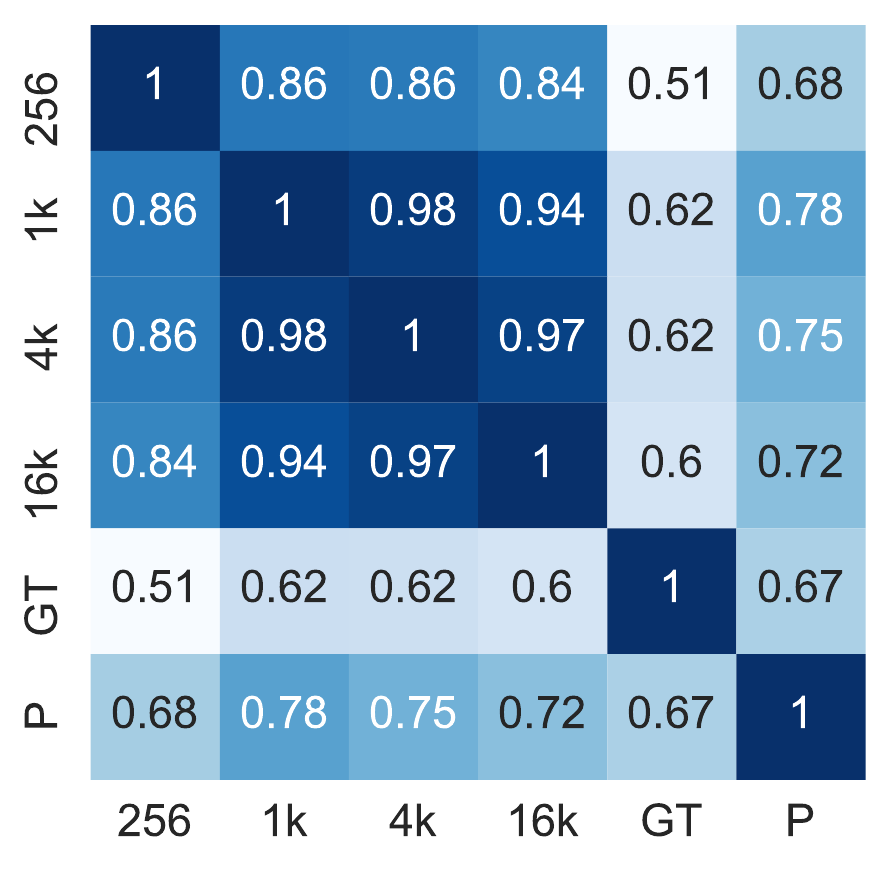}
        \caption{NAS-Bench-101}
    \end{subfigure}
    \begin{subfigure}{0.245\textwidth}
        \centering
        \includegraphics[width=\textwidth]{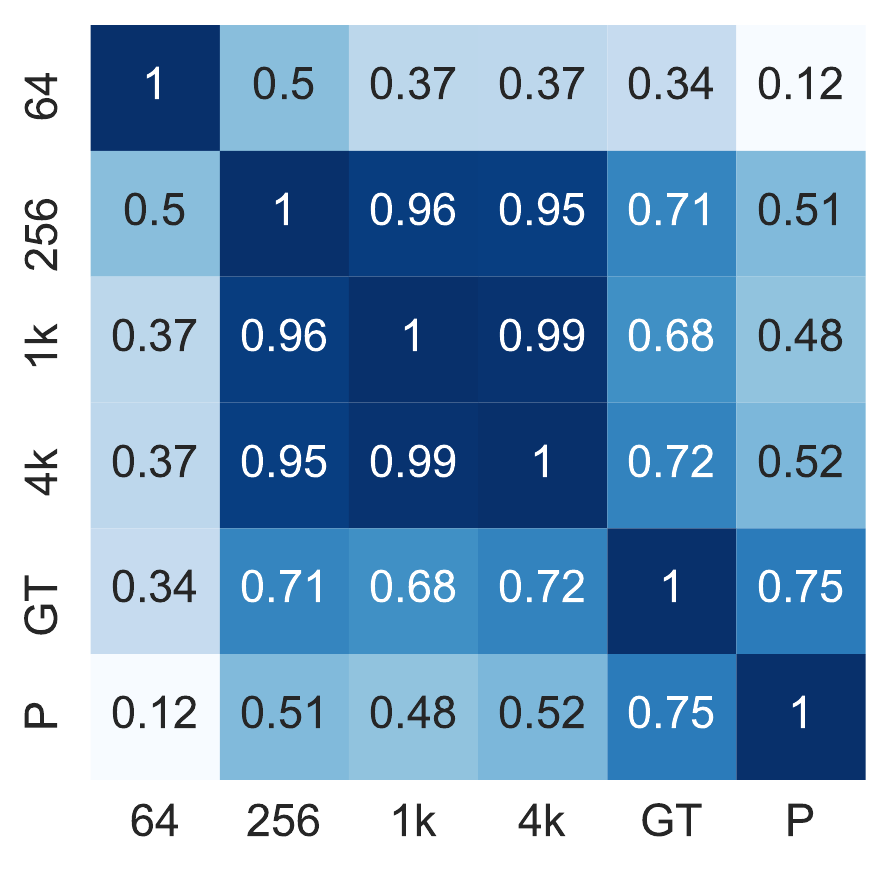}
        \caption{NAS-Bench-201}
    \end{subfigure}
    \begin{subfigure}{0.245\textwidth}
        \centering
        \includegraphics[width=\textwidth]{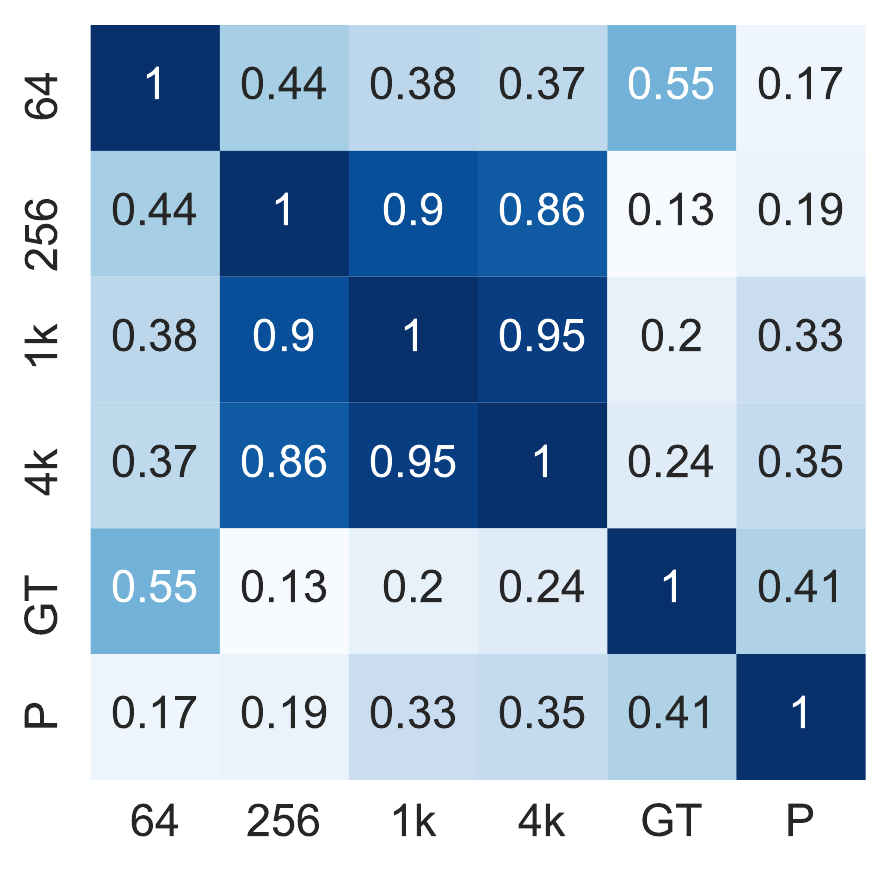}
        \caption{DARTS-CIFAR10}
    \end{subfigure}
    \begin{subfigure}{0.245\textwidth}
        \centering
        \includegraphics[width=\textwidth]{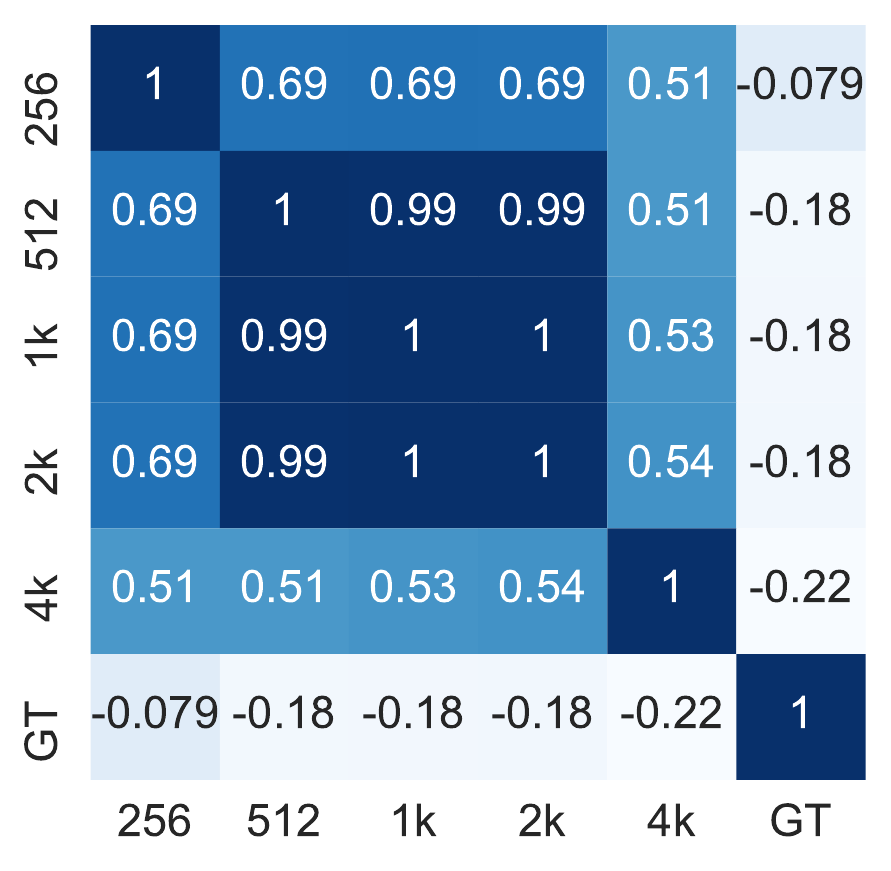}
        \caption{DARTS-PTB}
    \end{subfigure}
    \caption{Mutual correlation among the ranking produced by weight sharing under different budgets (number of epochs), number of parameters (P) and ground truth performance (GT).}
    \label{fig:correlation-mutual}
\end{figure}

\noindent \textbf{Weight sharing imposes a consistent bias.} One may attribute the failure of weight sharing to the instability of random-path gradient descent, bad hyper-parameters. But according to our study, that is not the case. As shown in \autoref{fig:correlation-mutual}, despite trained with different epochs, supernets are highly-agreeable when ranking architectures, even though they can be unanimously wrong (e.g., negative GT-correlation on DARTS-PTB). Similar consistency are also seen between different datasets, hyper-parameters and training tricks (\autoref{sec:supernet-training-settings}). The bias imposed by weight sharing seems consistent and lies in the intrinsics of weight sharing itself. 

\begin{wrapfigure}{l}{0.71\textwidth}
    \vspace{-1em}
    \centering
    \begin{subfigure}{0.23\textwidth}
        \centering
        \includegraphics[width=\textwidth]{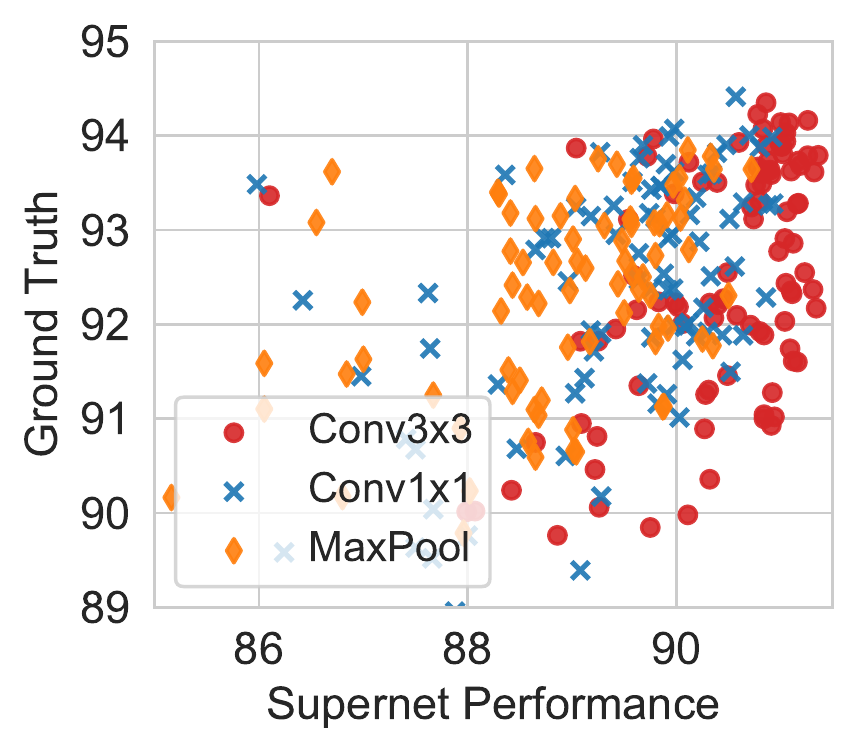}
        \caption{NB-101 (Node 1)}
    \end{subfigure}
    \begin{subfigure}{0.23\textwidth}
        \centering
        \includegraphics[width=\textwidth]{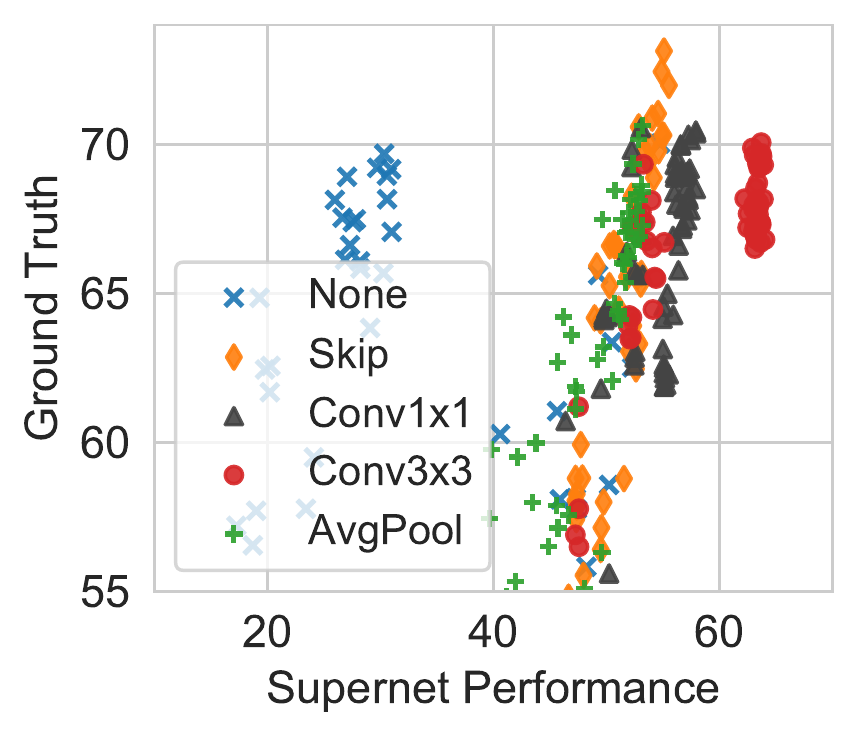}
        \caption{NB-201 (Edge 0-3)}
    \end{subfigure}
    \begin{subfigure}{0.23\textwidth}
        \centering
        \includegraphics[width=\textwidth]{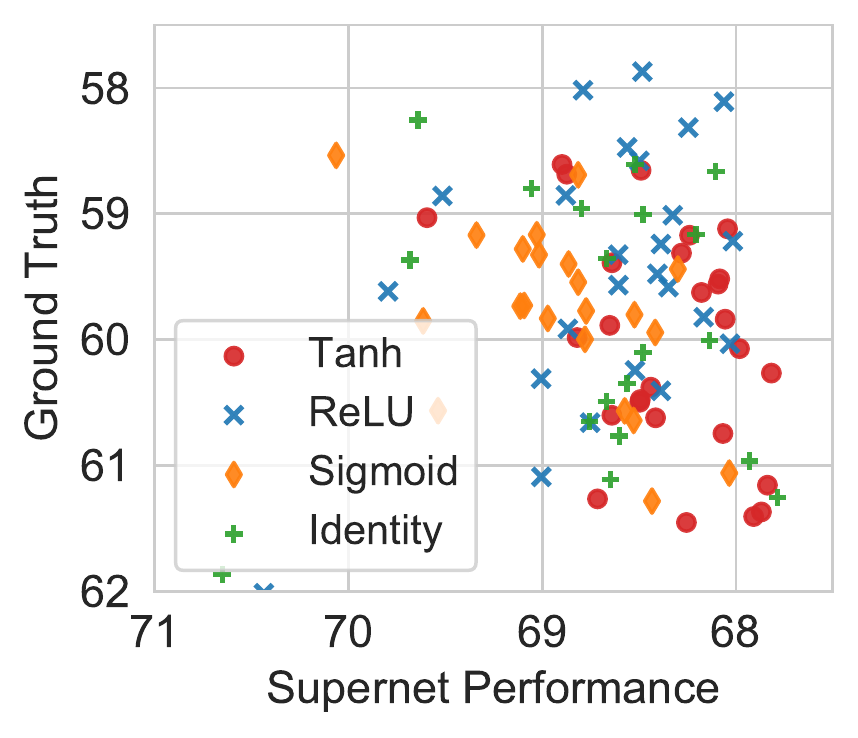}
        \caption{PTB (2nd activation)}
    \end{subfigure}
    \caption{Relationship between supernet performance and ground truth performance, grouped by decision on a specific operator, as stated in parenthesis. The most biased candidate is highlighted.}
    \label{fig:bias}
\end{wrapfigure}

\noindent \textbf{Weight sharing favors big models.} Architectures with large number of parameters (or FLOPs) usually have better ground truth accuracy~\citep{tan2019mnasnet,tan2019efficientnet}. As a result, for a search space spanning both very small and very large models (e.g., 1M to 13M parameters for NAS-Bench-101, 0.1M to 1.3M for NAS-Bench-201), model size can be a fairly good performance estimator baseline. As shown in \autoref{fig:correlation-mutual}, the correlation between ground truth performance and parameter size on NAS-Bench-101 can be as high as 0.67, outperforming all the attempts from weight sharing to estimate a good rank. Moreover, We find that weight sharing performs better in search spaces with higher GT-Param correlations. Sometimes, weight sharing even has a higher correlation with model size than its correlation with ground truth, in which case, weight sharing learns to find big models, rather than good-performing models.

\noindent \textbf{Weight sharing is biased towards certain operators.} We support this claim by reviewing the architecture distribution grouped by decisions of operators. Some typical cases are shown in \autoref{fig:bias}. For NAS-Bench-101, Conv3x3 is clearly preferred by weight sharing. Fortunately, architectures with Conv3x3 indeed have relatively good performance and that is why top-$k$ on NAS-Bench-101 is fairly good. On NAS-Bench-201, weight sharing still favors Conv3x3 as the operator connecting input node to output node of a cell, but almost all architectures with top ground truth accuracy choose the skip connection instead. This accounts for the fact that top-10 is almost no better than top-1 in \autoref{fig:top10-perf}. Similarly, on DARTS-PTB, Tanh, the activation function which is favored by weight sharing, turns out to be a worse option.

\section{Capacity of Weight Sharing}

\begin{wrapfigure}{r}{0.68\textwidth}
    \vspace{-1em}
    \centering
    \begin{subfigure}{0.33\textwidth}
        \centering
        \includegraphics[width=\textwidth]{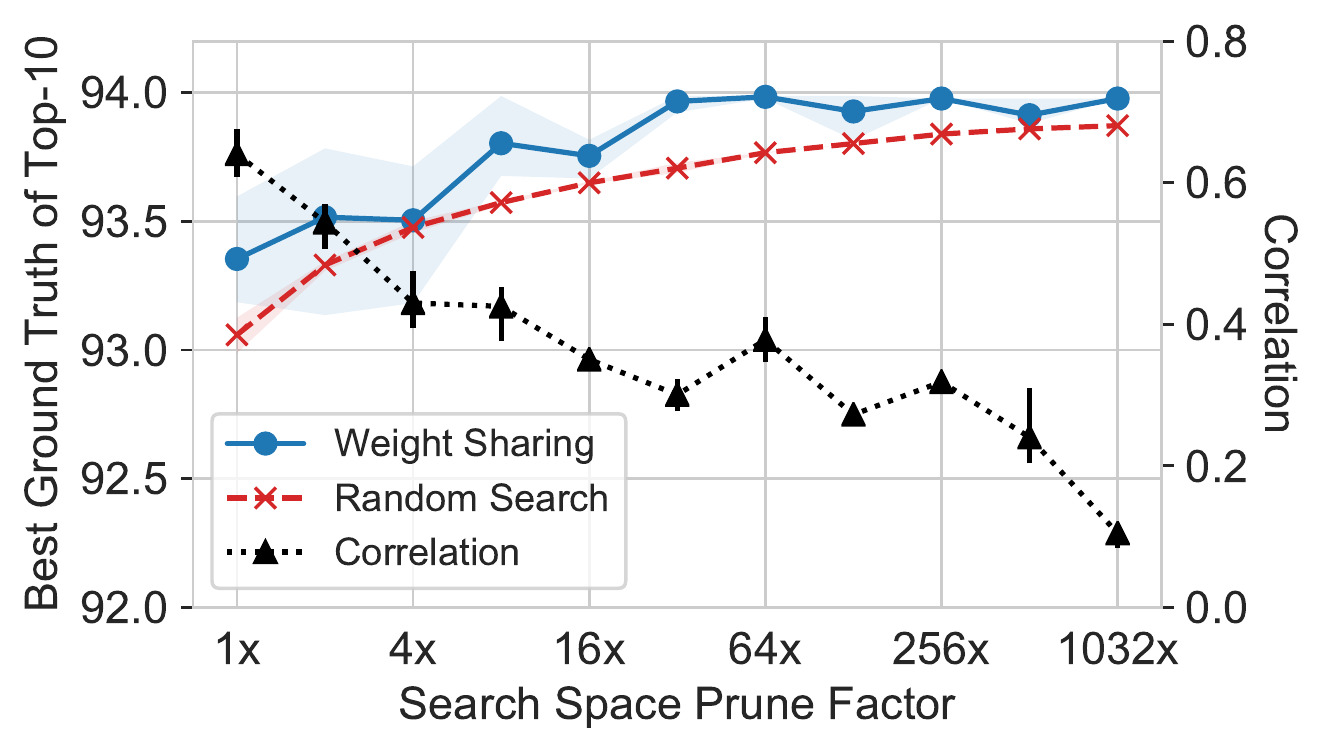}
        \caption{NAS-Bench-101}
    \end{subfigure}
    \begin{subfigure}{0.33\textwidth}
        \centering
        \includegraphics[width=\textwidth]{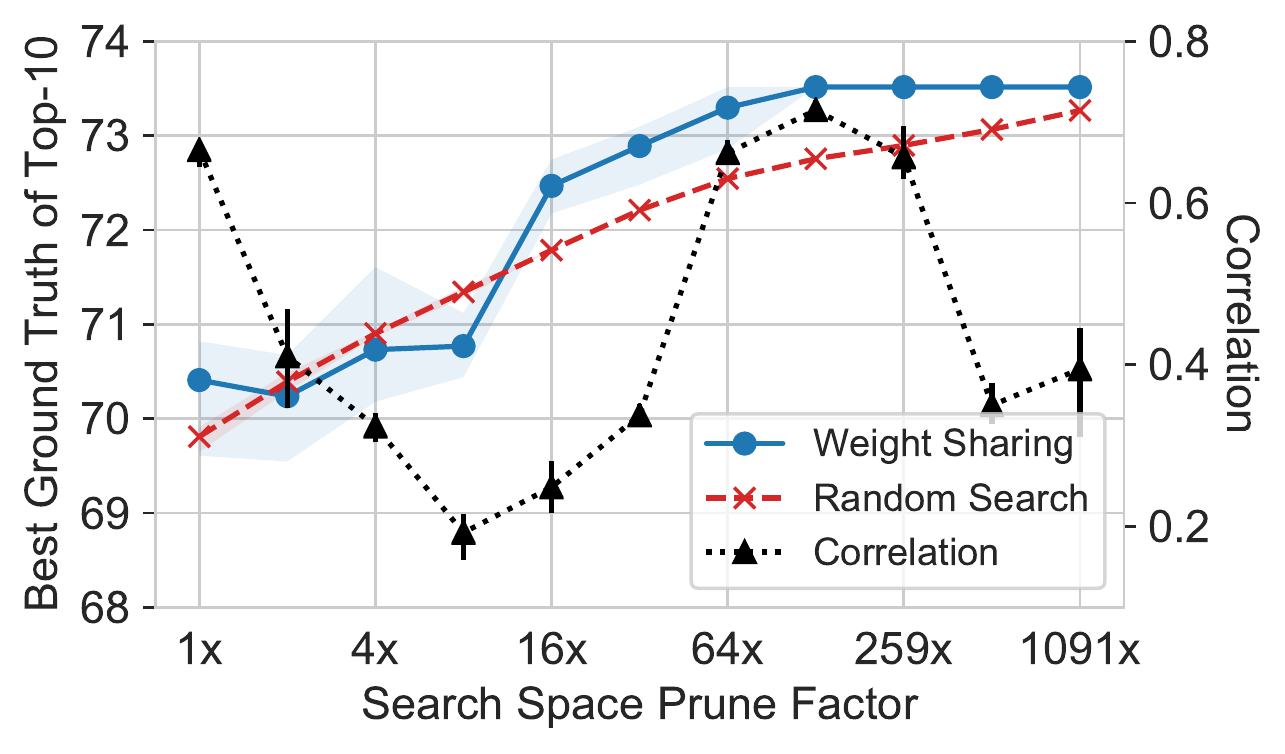}
        \caption{NAS-Bench-201}
    \end{subfigure}
    \caption{Comparison between random search and weight-sharing NAS on best-architecture subsets of search spaces. When search space is $k$ times smaller than the original one, we show top-10 accuracy (blue and red lines) for weight sharing and random search, respectively, and correlation in black lines. The error-bars are the min and max across 3 different runs.}
    \label{fig:prune-best}
\end{wrapfigure}

Since identifying the top architectures with weight sharing is difficult, previous works resort to weight sharing as a method to prune the worst, hoping to improve the performance distribution of search space, thus increase likelihood of finding the best. On search space weight sharing still works reasonably well (NAS-Bench-101 and 201), we ask the following question: is weight sharing still effective on a pruned search space? To answer the question, we test the capacity of weight sharing, after which prune factor NAS with weight sharing does not gain more benefits.

In particular, we run experiments on sub-search-space of NAS-Bench-101 and 201, by keeping a subset of architectures with the best ground truth performance (detailed setups in \autoref{sec:experiment-setup}). We vary the subset size and plot the figures shown in \autoref{fig:prune-best}. On NAS-Bench-101, we can see the advantage of weight sharing against random search becomes less obvious as the correlation goes down. This aligns with the intuition because the best architectures have very close accuracy and are hard to distinguish. Interestingly, on NAS-Bench-201, the advantage of weight sharing against random search is neither steadily increasing, nor gradually diminishes. It seems that weight sharing is reaching its limit when the prune factor is 8x, at which point, the advantage of weight sharing compared to random search becomes very little. However, after this point there could still be opportunities for weight sharing, as the correlation grows up to 0.6 when the space is pruned 128 times smaller. We conjecture that this phenomenon is a result of certain operators biased by weight sharing being removed from search space (e.g., the bad-performing Conv3x3 on NAS-Bench-201).

\section{Conclusion}

In this paper, we give a comprehensive view of weight sharing in neural architecture search. Through our experiments, we have shown that weight sharing, though effective on some search spaces, can be limited on others. Weight sharing imposes a bias towards certain architectures and its capacity is limited to discriminate better architectures from worse. Our findings call for a better understanding of such biases and co-design of search space and weight sharing NAS. Meanwhile, instead of directly identifying the top, weight-sharing-guided search space pruning is shown to be a promising lead, and we hope this will inspire future work.

{
\small
\bibliography{iclr2021_conference}
\bibliographystyle{iclr2021_conference}
}

\appendix

\section{Experiment Setups}
\label{sec:experiment-setup}

\subsection{Search Spaces}
\label{sec:search-space}

We present a summary of 5 search spaces we used for our experiments in \autoref{tab:search-space-summary}. Below is a detailed description of the 5 search spaces and how we use weight sharing on these search spaces.

\paragraph{NAS-Bench-101.} To make diagnose of one-shot methods possible and fully leverage previous-computed benchmark results, we design a sub search space of NAS-Bench-101~\citep{ying2019nasbench101}, which is also referred in NAS-Bench-1Shot1~\citep{zela2020nasbench1shot1}. Our search space is similar to search space 3 in NAS-Bench-1Shot1, with subtle differences to fully align with NAS-Bench-101: (i) We use the same channel number configuration as NAS-Bench-101. (ii) We force the number of intermediate nodes in a cell to be exactly 5, to avoid introducing an extra zero-op. (iii) In NAS-Bench-101, all intermediate nodes are concatenated to the output node of a cell, after which input node (after 1x1 conv if necessary) is added to the output. The channel number of intermediate nodes is the output channel divided by the number of intermediate nodes (instead of all incoming nodes). Instead of introducing dynamic convolutions, we forbid the connection from input to output, force the output node to concatenate 2 of the intermediate 5 nodes. (iv) We do not deduplicate isomorphic cells. In the end, we end up with 95985 different cells, and 42228 deduplicate isomorphic ones. Following the guidelines recommended by NAS-Bench-101, we split the original CIFAR-10 training set into 40k for training and 10k for validation. The original test set is used for testing.

\paragraph{NAS-Bench-201.} NAS-Bench-201~\citep{dong2020nasbench201} constructs its cell with 4 nodes connected with 6 edges, with an operator choosing from 5 candidate operators on each edge. In NAS-Bench-201, 341 out of 15625 are invalid, where input node of the cell is completely not connected to output node, we simply devise a filter to avoid those invalid cells from being chosen. Though the original NAS-Bench-201 is given across three different scaled image classification datasets, we use CIFAR-100 only as architecture rank on three datasets are reported to be highly correlated and the other two datasets have already been used on other search spaces.

\paragraph{DARTS-CIFAR10.} The search space on CIFAR-10 dataset adopted in DARTS~\citep{liu2018darts} is very similar to the ones used by \citet{pham2018efficient,zoph2018learning}, except that all nodes in a cell are concatenated into the output and there are subtle differences in convolution with stride = 2 in reduction cell. Previous works handle the search space differently. For example, DARTS expands the combinations of operators and inputs, thus the supernet is 3 -- 4 times larger than that constructed following ENAS~\citep{pham2018efficient}. In our paper, we use the straight-forward supernet-building which is proposed in ENAS.

\paragraph{ProxylessNAS.} We follow ProxylessNAS-GPU~\citep{cai2018proxylessnas} setting, with 1.35 width multiplier. During training, all architectures are equally likely to be sampled. In evaluation, we only consider architectures that fit in mobile settings and filter out the architectures with FLOPs greater than 600M.

\paragraph{DARTS-PTB.} We use the RNN cell search space consisting of 8 choices of activation function and choices of predecessors, following \citet{liu2018darts}. We use the official train, valid and test split provided by Penn Treebank.

We present a summary of all search spaces in \autoref{tab:search-space-summary}.

\begin{table}[htbp]
    \caption{Summary of search spaces.}
    \label{tab:search-space-summary}
    \begin{adjustbox}{width=\textwidth}
    \begin{tabular}{c|c|c|c|c}
        \hline
        & Valid space size & Dataset & Task & Space type \\
        \hline
        NAS-Bench-101 & 95985 & CIFAR-10 & Image classification & Cell-wise~\cite{zoph2018learning} \\
        NAS-Bench-201 & 15284 & CIFAR-100 & Image classification & Cell-wise \\
        DARTS-CIFAR10 & $2.76 \cdot 10^{20}$ & CIFAR-10 & Image classification & Cell-wise \\
        ProxylessNAS & $2.22 \cdot 10^{17}$ & ImageNet & Image classification & Chain-wise~\cite{zoph2016neural} \\
        DARTS-PTB & $2.64 \cdot 10^9$ & Penn Treebank & Language modelling & Cell-wise \\
        \hline
    \end{tabular}
    \end{adjustbox}
\end{table}

\subsection{Evaluation Methodology}

\label{sec:evaluation-methodology}

\paragraph{Single-path one shot.} In single-path one shot, an architecture is sampled per mini-batch with the objective of minimizing the expectation of loss, i.e.,

\begin{equation}
    \label{eq:target-eq}
    \E_{\alpha \sim \Arch} [ \Ls_{val} (\Net(\phi, \alpha)) ]
\end{equation}

where $\Arch$ is the architecture search space, $\phi$ is the supernet weights, and $\Net$ is the corresponding architecture using weights from supernet. When evaluating a submodel, we first extract it from supernet inheriting supernet weights, calibrate Batch-Norm statistics~\citep{guo2019single,bender18understanding,pourchot2020share,yu2020train}, and then evaluate it on the full validation set (top-1 accuracy for image classification task and perplexity for language modelling). When 
calculating the metrics (including correlation and top-k), we sample and evaluate 1k architectures for NAS-Bench series, 100 for DARTS series, and 20 for ProxylessNAS.

\paragraph{Hyper-parameter settings.} We list important hyper-parameters for supernet training in \autoref{tab:hparams-supernet}. We run the training with 1000 epochs if not otherwise specified.

\begin{table}[h]
    \centering
    \caption{Summary of hyper-parameters used in supernet training. BN calib batch size / steps~\citep{bender18understanding} are the configuration of batch-norm calibration before evaluation of each architecture.}
    \label{tab:hparams-supernet}
    \begin{tabular}{c|c|c|c|c|c}
    \hline
    & NB-101 & NB-201 & DARTS-C & Proxyless & DARTS-PTB \\
    \hline
    Batch size & 256 & 256 & 448 & 1024 & 256 \\
    Initial LR & 0.1 & 0.05 & 0.1 & 0.2 & 40 \\
    LR Decay   & Cosine & Cosine & Cosine & Cosine & Fixed \\
    Weight decay & 0.0001 & 0.0005 & 0.0003 & 0.0005 & $8 \cdot 10^7$ \\
    Gradient clip & 5 & 5 & 5 & 0 & 0.1 \\
    BN calib batch size & 400 & 400 & 200 & 400 & - \\
    BN calib steps      & 25  & 25  & 100 & 25  & 0 \\
    \# GPUs       & 1 & 1 & 4 & 8 & 1 \\
    \hline
\end{tabular}
\end{table}

For ground truth training, we completely followed the settings mentioned in the original paper, except those which are tuned to fit into our GPU capacity (batch size = 96 for DARTS-CIFAR10, and 1024 for ProxylessNAS).

\paragraph{Metrics.} Following previous works, we use the following metrics to assess the effectiveness of weight sharing in NAS:

\begin{itemize}[leftmargin=0.3in]
    \item {\bf Correlation}: Following \citet{yu2020train,pourchot2020share,zhang2020deeper,li2019random}, we show the correlation between supernet performance and ground truth performance. We will use Spearman correlation~\citep{spearman1961proof} in this paper. A higher correlation means better consistency. Correlations are calculated on a randomly sampled subset (1k for NAS-Bench series, 100 for DARTS series, 20 for ProxylessNAS).
    \item {\bf Top-$k$}: We randomly sample 1k architectures, choose top-$k$ based on their performance evaluated on supernet, and obtain their ground truth performance. We then use report test performance of the architecture with the best ground truth performance on validation set. 
\end{itemize}

\paragraph{Search space pruning.} To keep the best performing subset of search space, we adopt the method to keep the architectures with best ground truths. In practice, the ground truths are usually not available and replaced with the estimation by weight sharing methods~\citep{pham2018efficient,you2020greedynas}. Note that we did not prune the search space at operator level, as previously done in \citet{chen2019pdarts,hu2020anglebased}.

\section{Performance Benchmark}
\label{sec:benchmark}

\begin{figure}[htbp]
    \centering
    \begin{subfigure}{.3\textwidth}
        \centering
        \includegraphics[width=\textwidth]{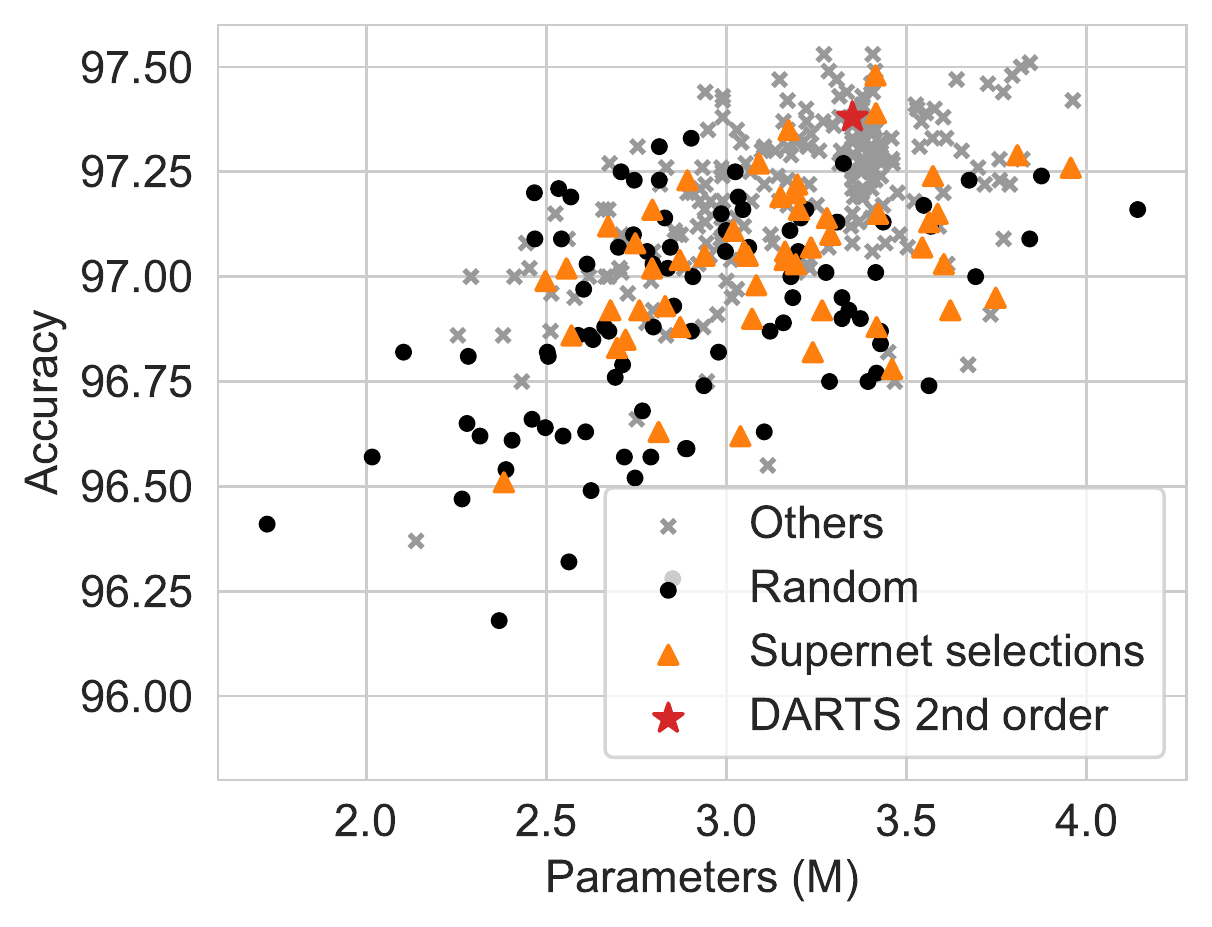}
        \caption{DARTS-CIFAR10}
    \end{subfigure}
    \begin{subfigure}{.32\textwidth}
        \centering
        \includegraphics[width=\textwidth]{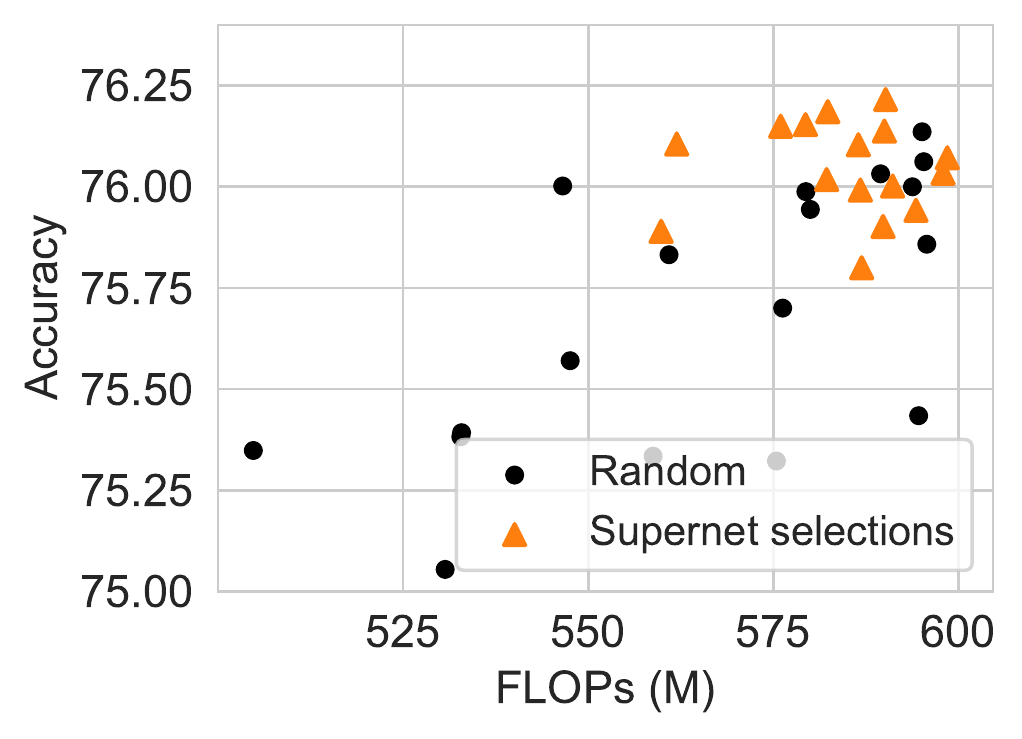}
        \caption{ProxylessNAS}
    \end{subfigure}
    \begin{subfigure}{.28\textwidth}
        \centering
        \includegraphics[width=\textwidth]{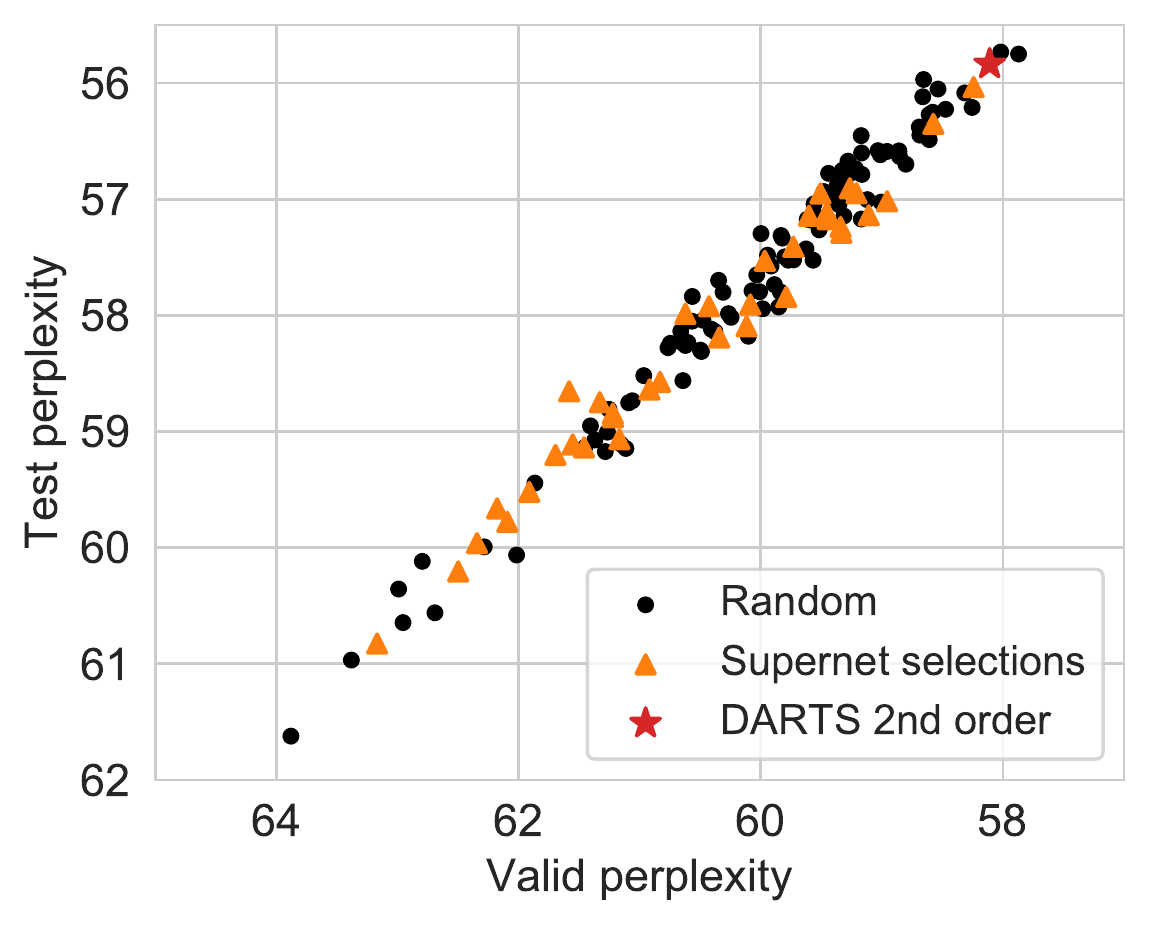}
        \caption{DARTS-PTB}
    \end{subfigure}
    \caption{Efficiency-latency distribution of our sampled ground-truths. For DARTS-PTB, we show the relationship between validation perplexity and test perplexity because model sizes are same for all architectures.}
    \label{fig:performance-benchmark}
\end{figure}

For the purposes of conducting multiple experiments mentioned in our paper, we trained 377 architectures on DARTS-CIFAR10, 36 architectures on ProxylessNAS and 137 architectures on DARTS-PTB, independently from scratch. We show the distribution in \autoref{fig:performance-benchmark}, where each point corresponds to an architecture, and publish these results for the usage of future researches.

\section{Effect of Supernet Training Tricks}
\label{sec:supernet-training-settings}

\begin{figure}[htbp]
    \centering
    \begin{subfigure}{.48\textwidth}
        \centering
        \includegraphics[width=\textwidth]{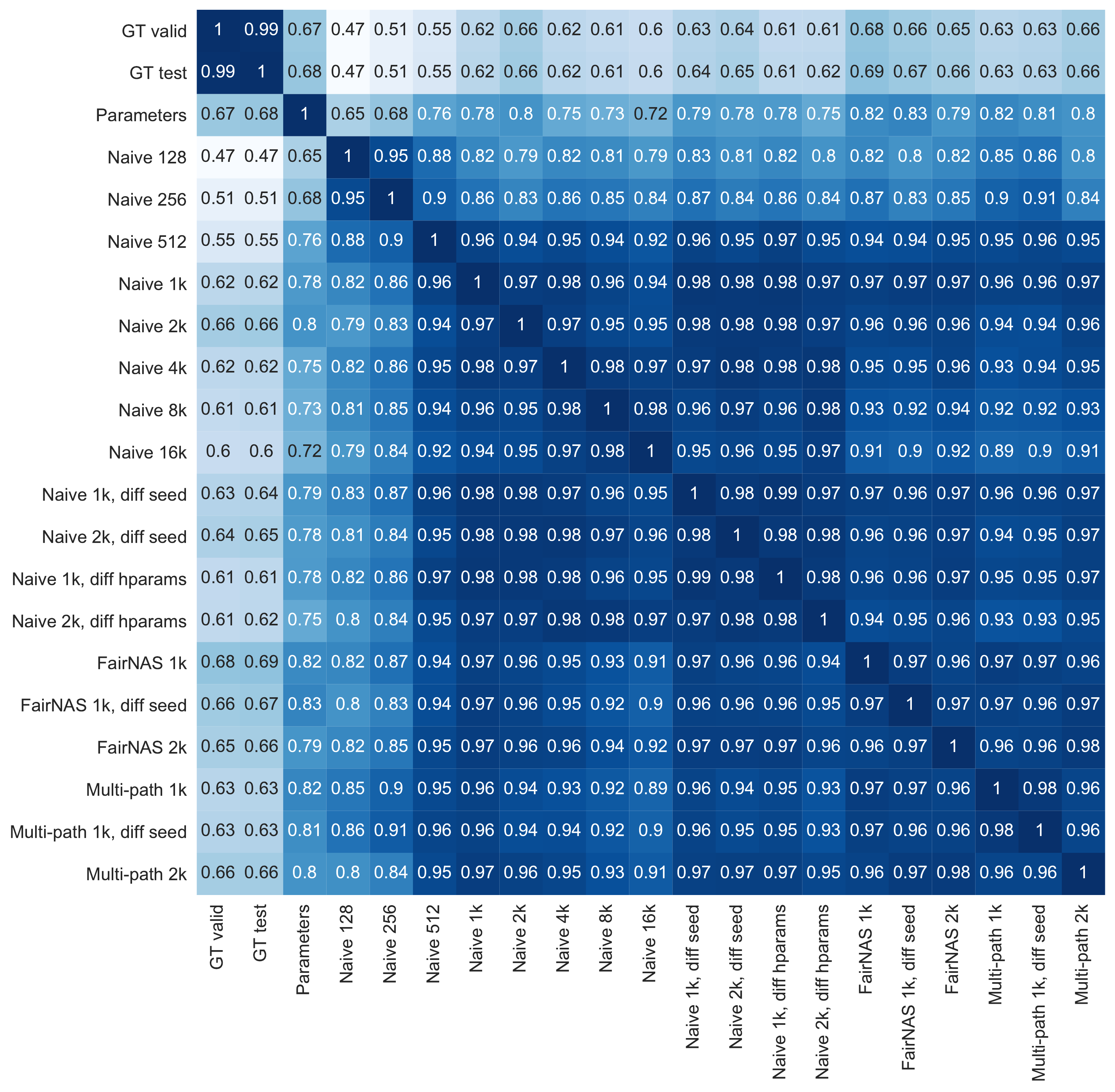}
        \caption{NAS-Bench-101}
    \end{subfigure}
    \begin{subfigure}{.48\textwidth}
        \centering
        \includegraphics[width=\textwidth]{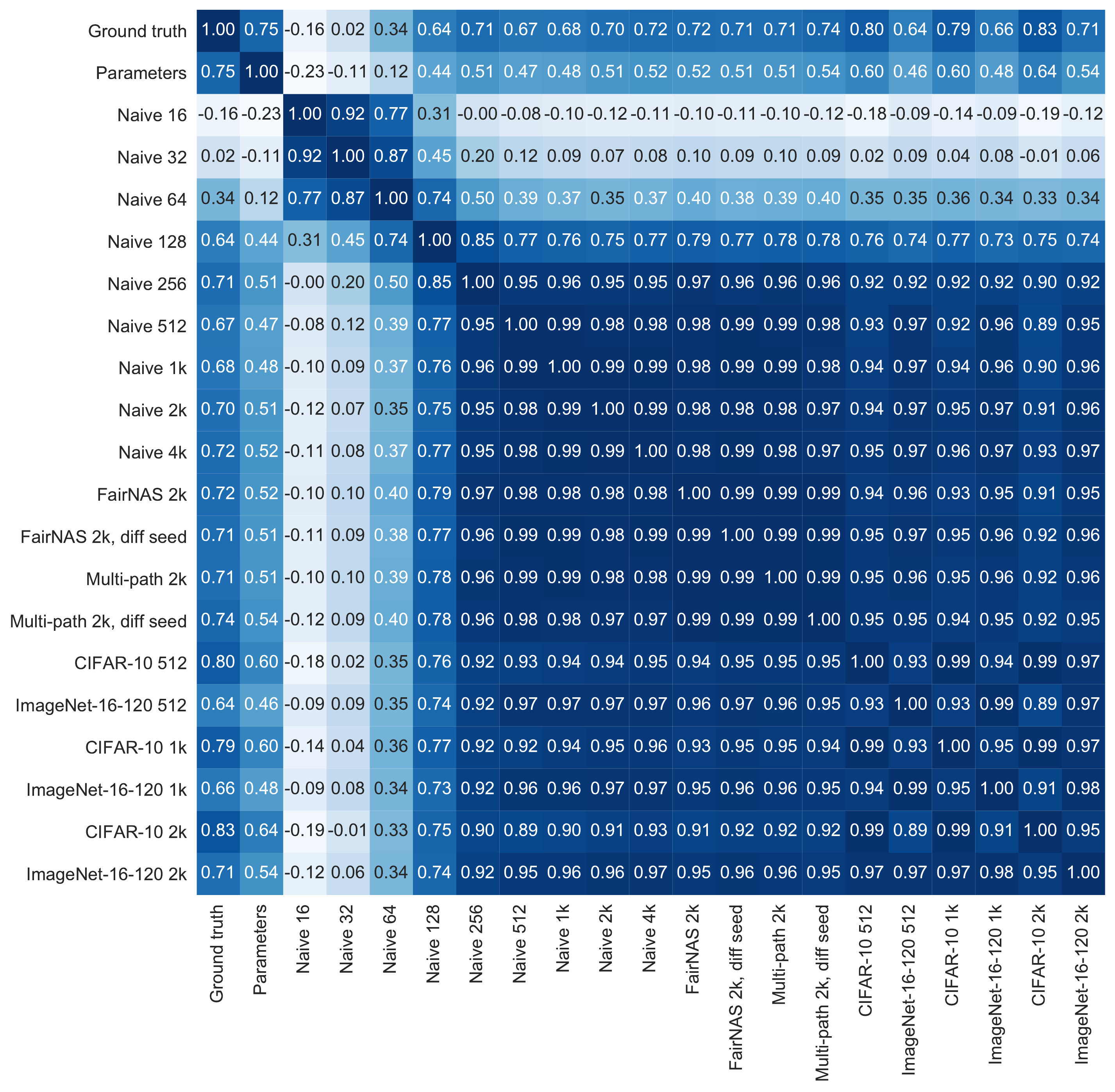}
        \caption{NAS-Bench-201}
    \end{subfigure}
    \begin{subfigure}{.3\textwidth}
        \centering
        \includegraphics[width=\textwidth]{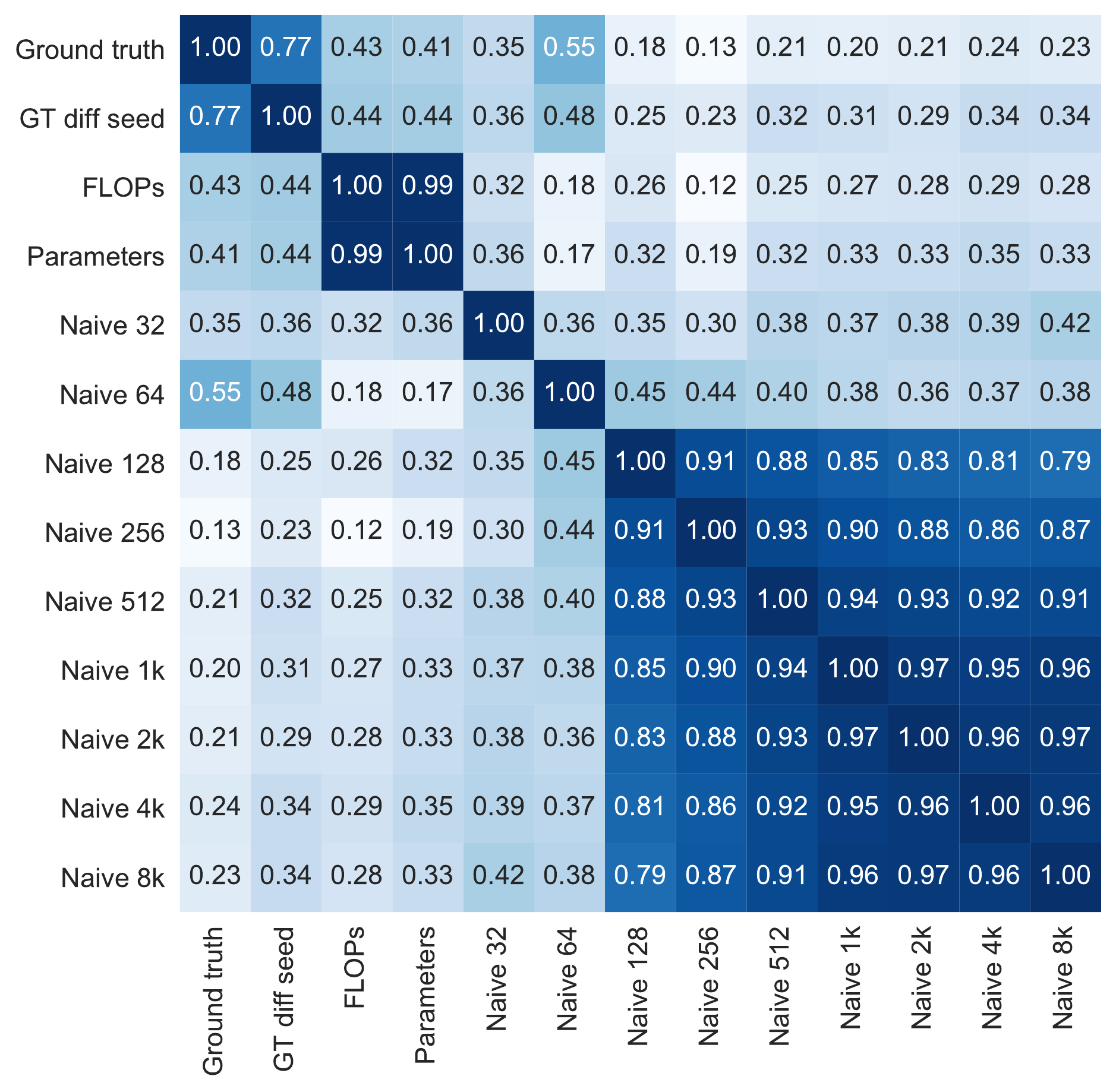}
        \caption{DARTS-CIFAR10}
    \end{subfigure}
    \begin{subfigure}{.28\textwidth}
        \centering
        \includegraphics[width=\textwidth]{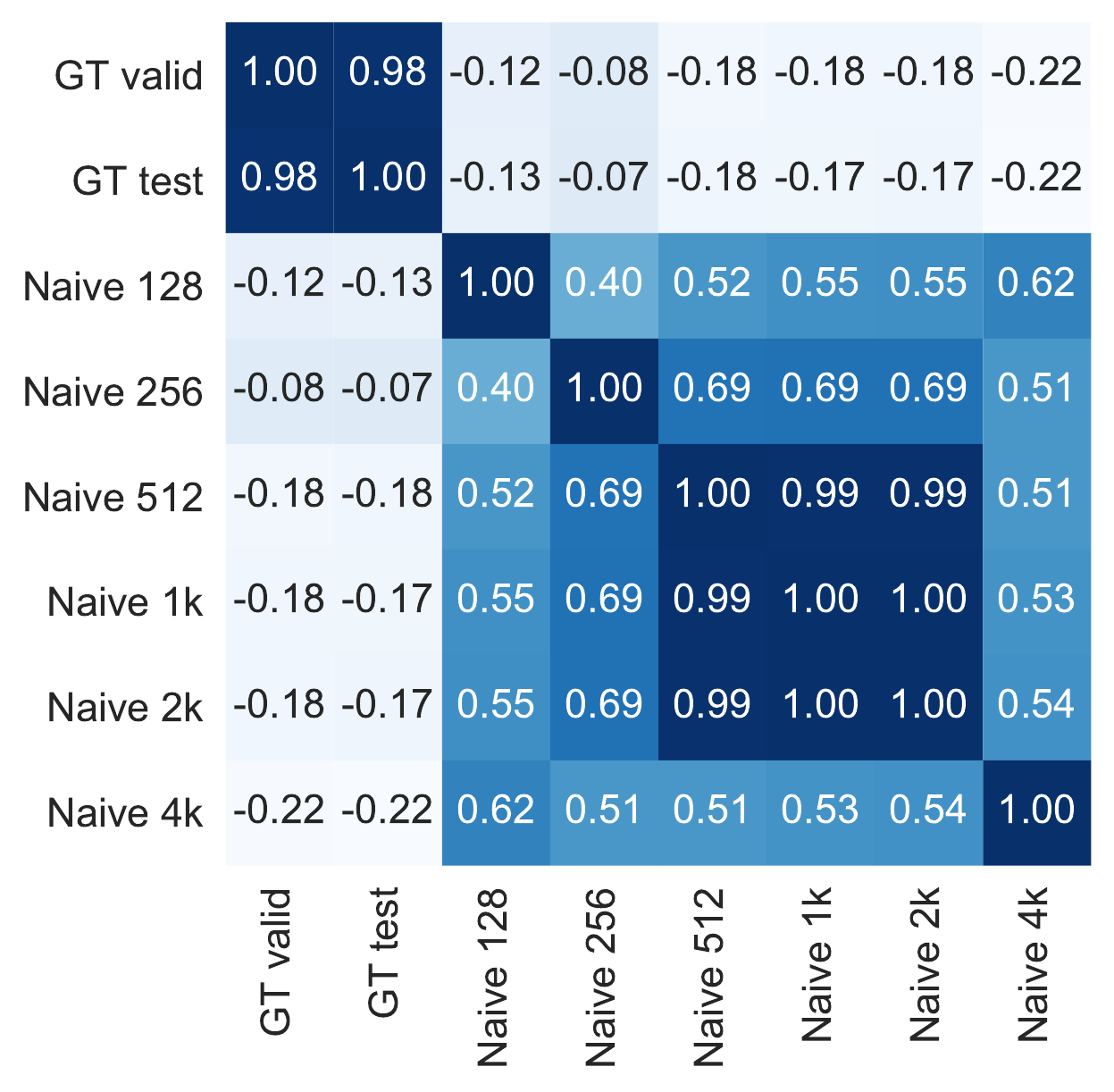}
        \caption{DARTS-PTB}
    \end{subfigure}
    \begin{subfigure}{.28\textwidth}
        \centering
        \includegraphics[width=\textwidth]{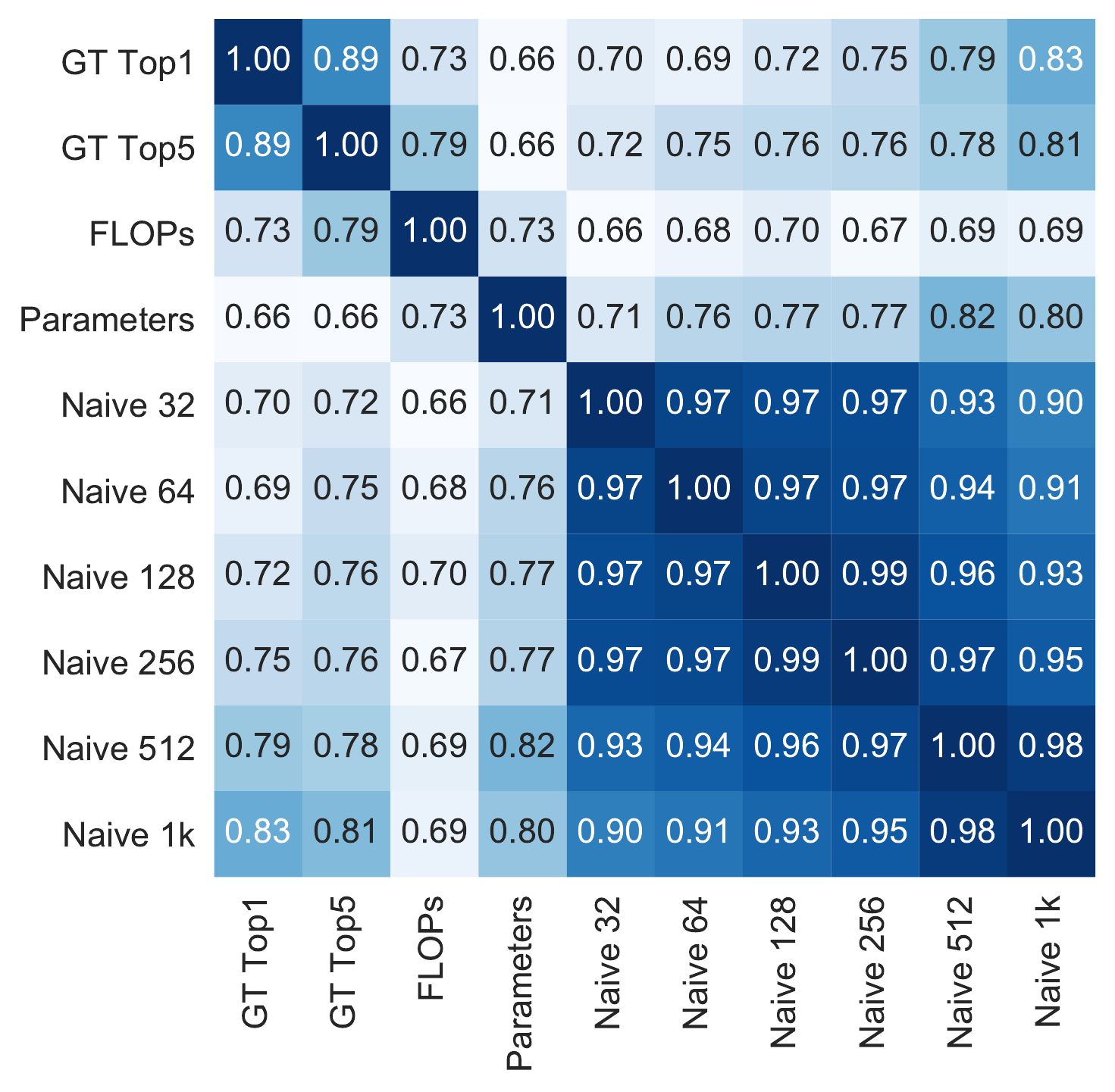}
        \caption{ProxylessNAS}
    \end{subfigure}
    \caption{Mutual correlations of all trained supernets and ground truths on various search spaces.}
    \label{fig:all-correlation}
\end{figure}

Most of the experiments in this paper are run with the training approach and hyper-parameters listed in \autoref{sec:experiment-setup}. However, recent works have proposed many variations based on SPOS~\citep{yang2019cars,pham2018efficient,you2020greedynas,chu2019fairnas}. Even if we faithfully follow SPOS to train a supernet, there are still quite a few hyper-parameters to tune. We test the impact of different settings and show the results in \autoref{tab:supernetstab} and \autoref{fig:all-correlation}. Judging from the results, there is no evidence that applying any of the tricks is beneficial. Meanwhile, almost all weight sharing methods are correlated with each other, implying that the bias of weight sharing is consistent.

\begin{table}[htbp]
    \centering
    \caption{Comparison of correlation with ground truth (Corr-GT), top-10, and correlation with baseline (Corr-BL) after varying training settings and applying training tricks.}
    \label{tab:supernetstab}
\begin{tabular}{c|c|c|c}
    \hline
    Group & Corr-BL & Corr-GT $\uparrow$ ($\Delta$) & Top-10 $\uparrow$ ($\Delta$) \\
    \hline
    NAS-Bench-101 (1k epochs)       &   1.000 &                0.618 &                93.45 \\
    FairNAS~\citep{chu2019fairnas}   &   0.967 &  0.683 (\textcolor{blue}{+0.064}) &  93.64 (\textcolor{blue}{+0.19})  \\
    Multi-Path~\citep{pham2018efficient} &   0.963 &  0.626 (\textcolor{blue}{+0.007}) &  93.87 (\textcolor{blue}{+0.42}) \\
    LR=0.05, WD=$3 \cdot 10^4$      &   0.982 &   0.608 (\textcolor{red}{-0.010}) &   93.13 (\textcolor{red}{-0.32}) \\
    Different Seed                  &   0.982 &  0.630 (\textcolor{blue}{+0.012}) &  93.59 (\textcolor{blue}{+0.14}) \\
    \hline
    NAS-Bench-101 (2k epochs)       &   1.000 &                0.659 &                93.82  \\
    FairNAS                         &   0.964 &   0.652 (\textcolor{red}{-0.007}) &   93.41 (\textcolor{red}{-0.41}) \\
    Multi-Path                      &   0.963 &   0.656 (\textcolor{red}{-0.002}) &   93.64 (\textcolor{red}{-0.18}) \\
    LR=0.05, WD=$3 \cdot 10^4$      &   0.970 &   0.611 (\textcolor{red}{-0.048}) &   93.45 (\textcolor{red}{-0.37}) \\
    Different Seed                  &   0.979 &   0.644 (\textcolor{red}{-0.014}) &   93.45 (\textcolor{red}{-0.37}) \\
    \hline
    NAS-Bench-201 (2k epochs)       &   1.000 &                0.701 &                  69.66 \\
    FairNAS &   0.978 &  0.716 (\textcolor{blue}{+0.015}) &  69.87 (\textcolor{blue}{+0.21}) \\
    Multi-Path &   0.984 &  0.706 (\textcolor{blue}{+0.005}) &  70.05 (\textcolor{blue}{+0.39})  \\
    CIFAR-10 &   0.915 &  0.831 (\textcolor{blue}{+0.130}) &  69.87 (\textcolor{blue}{+0.21}) \\
    ImageNet-16-120 &   0.962 &  0.711 (\textcolor{blue}{+0.010}) &  69.72 (\textcolor{blue}{+0.06}) \\
    \hline
\end{tabular}

\end{table}

\section{Convergence of Supernet}
\label{sec:convergence}

\begin{figure}[t]
    \centering
    \begin{subfigure}{0.4\textwidth}
        \centering
        \includegraphics[width=\textwidth]{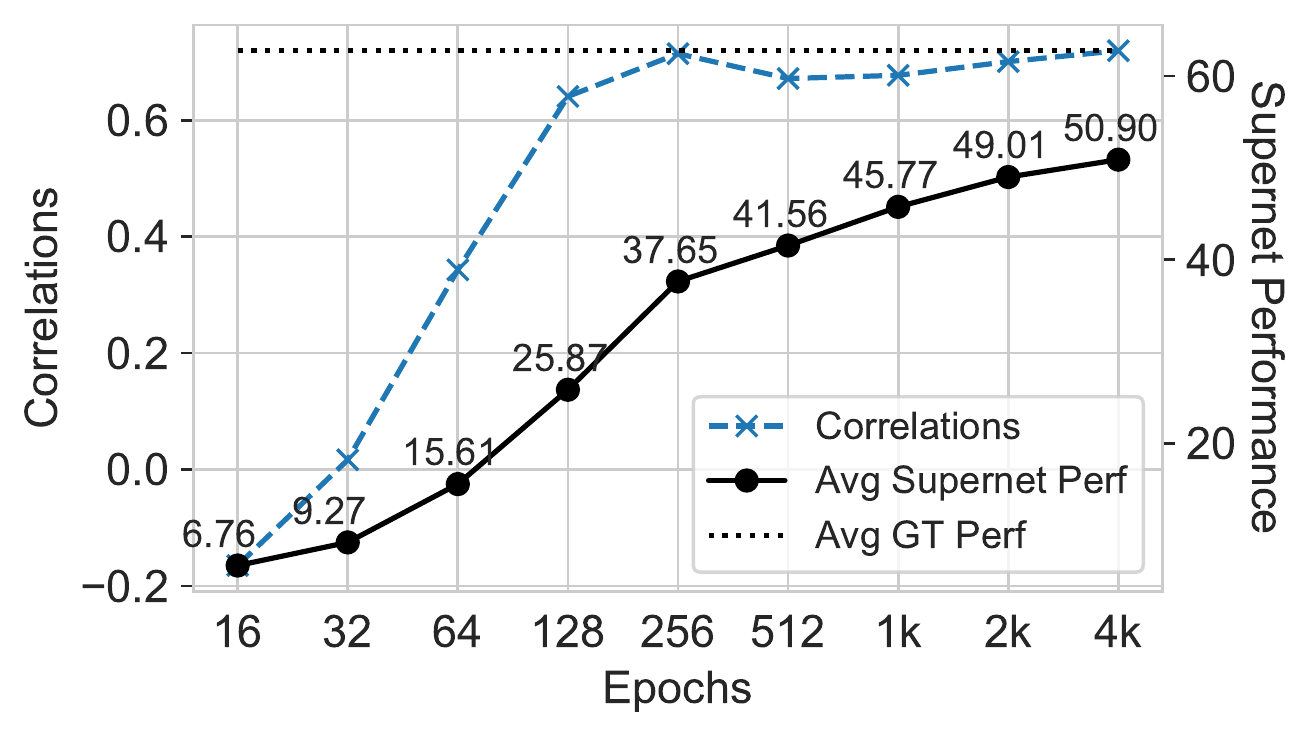}
        \caption{NAS-Bench-201}
    \end{subfigure}
    \begin{subfigure}{0.4\textwidth}
        \centering
        \includegraphics[width=\textwidth]{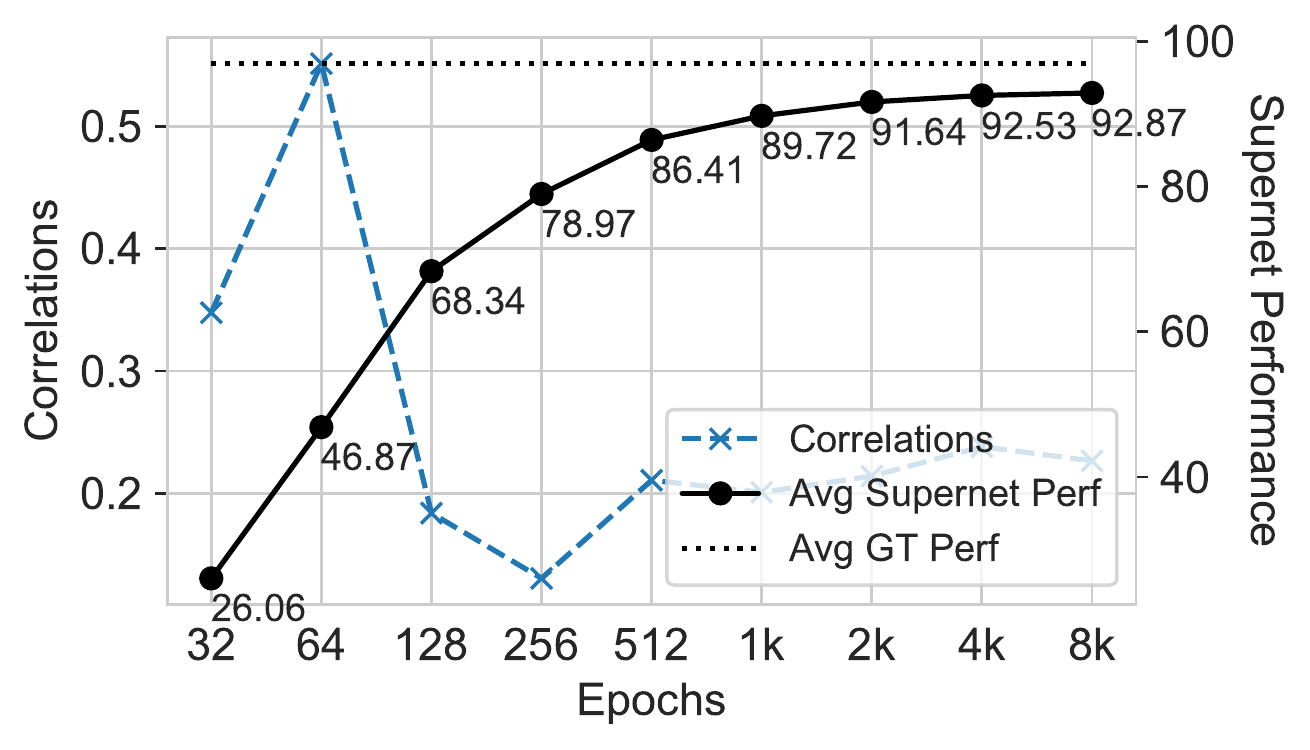}
        \caption{DARTS-CIFAR10}
    \end{subfigure}
    \begin{subfigure}{0.39\textwidth}
        \centering
        \includegraphics[width=\textwidth]{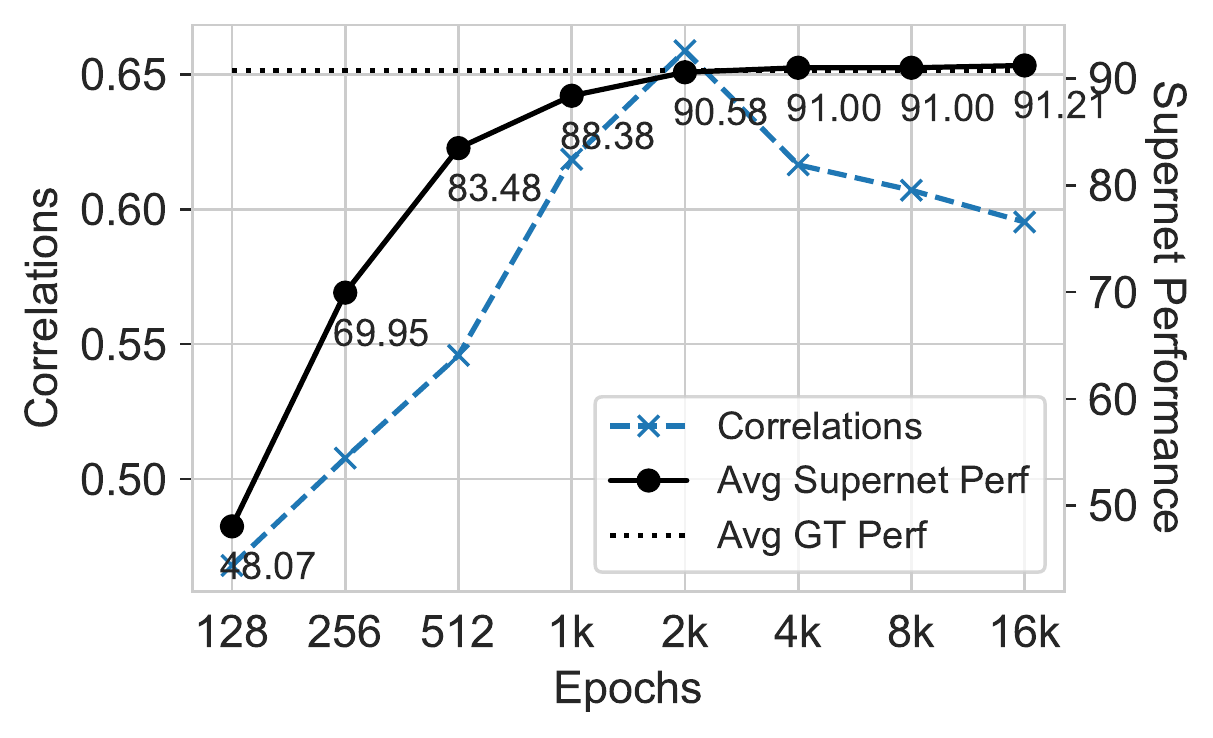}
        \caption{NAS-Bench-101}
    \end{subfigure}
    \begin{subfigure}{0.24\textwidth}
        \centering
        \includegraphics[width=\textwidth]{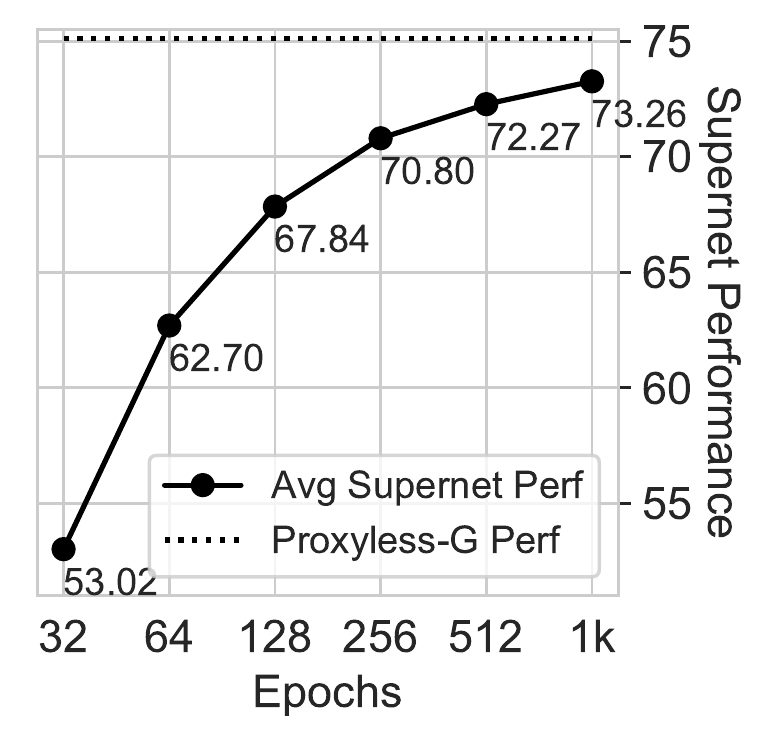}
        \caption{ProxylessNAS}
    \end{subfigure}
    \begin{subfigure}{0.35\textwidth}
        \centering
        \includegraphics[width=\textwidth]{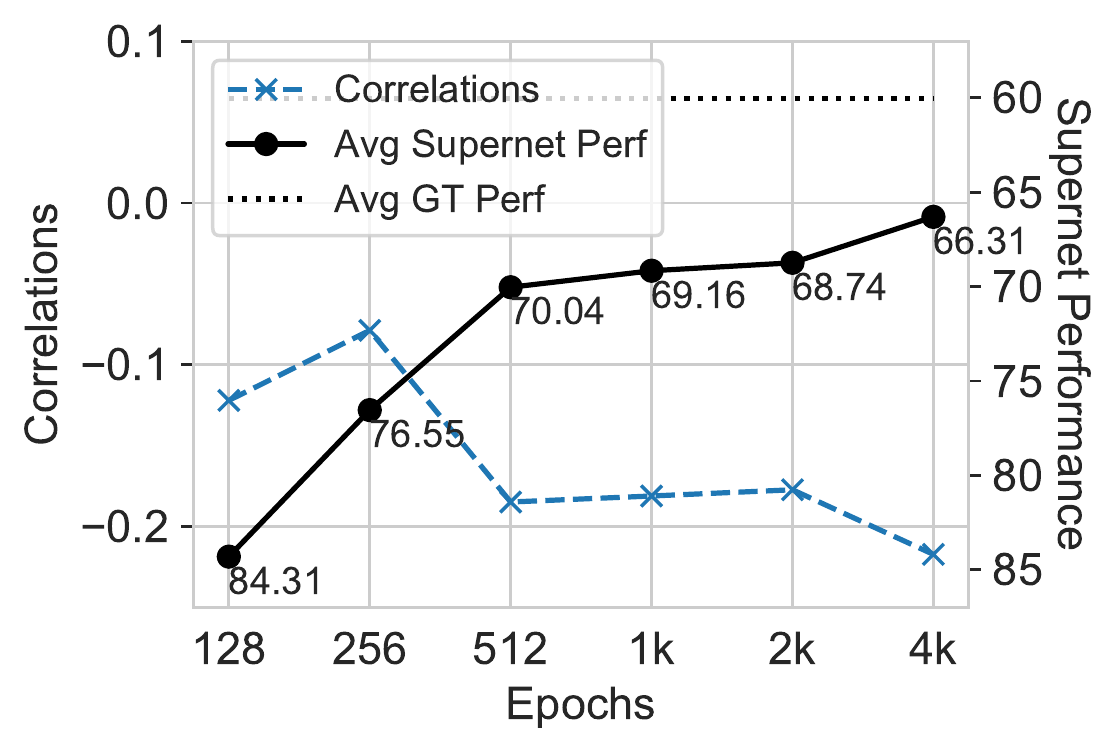}
        \caption{DARTS-PTB}
    \end{subfigure}
    \caption{Accuracy and correlation with respect to the number of epochs. The dotted line is the average ground truth performance, i.e., performance of architectures when they are trained independently. Correlation line on ProxylessNAS is missing because we could not get a fair number of ground truths, and the dotted line is the accuracy of ``Proxyless-GPU'' architecture.}
    \label{fig:correlation-error}
\end{figure}

Here, we discuss the convergence of supernet. We computed the average supernet performance, i.e., average accuracy of submodels evaluated on supernet. As shown in \autoref{fig:correlation-error}, there is a clear trend that extended training will boost supernet performance on all five search spaces and the performance is expected to continue to increase the training epochs. There is no sign of over-fitting. On NAS-Bench-101, supernet performance even surpasses average ground truth. Better average accuracy implies that the validation loss in \autoref{eq:target-eq} is optimized better. However, this does not mean the supernet is more capable of identifying top architectures, as the correlation sometimes goes lower.

\end{document}